\documentclass{article}

\usepackage{microtype}
\usepackage{graphicx}
\usepackage{subfigure}
\usepackage{booktabs} % for professional tables

\usepackage{hyperref}

\usepackage[accepted]{icml2024}

\usepackage{amsmath}
\usepackage{amssymb}
\usepackage{mathtools}
\usepackage{amsthm}
\usepackage{bm}
\usepackage{multirow}
\usepackage{subfigure}
\usepackage{array}
\usepackage{rotating}
\usepackage{pifont}
\usepackage{wrapfig}
\usepackage{caption} 
\usepackage{tablefootnote}
\usepackage{savefnmark}
\usepackage{rotating}
\usepackage{tikz}
\usepackage{dcolumn, array, booktabs}
\usetikzlibrary{tikzmark, calc}

\usepackage[capitalize,noabbrev]{cleveref}

\theoremstyle{plain}

\theoremstyle{definition}

\theoremstyle{remark}

\newcommand{\figref}[1]{Figure \ref{#1}}
\newcommand{\tabref}[1]{Table \ref{#1}}
\newcommand{\secref}[1]{Section \ref{#1}}
\newcommand{\equref}[1]{Equation (\ref{#1})}
\usepackage[disable,textsize=tiny]{todonotes}
\icmltitlerunning{Language Models are Super Mario: Absorbing Abilities from Homologous Models as a Free Lunch}
\begin{document}

\twocolumn[
\icmltitle{Language Models are Super Mario: \\
           Absorbing Abilities from Homologous Models as a Free Lunch}



\icmlsetsymbol{equal}{*}

\begin{icmlauthorlist}
\icmlauthor{Le Yu}{comp}
\icmlauthor{Bowen Yu}{comp}
\icmlauthor{Haiyang Yu}{comp}
\icmlauthor{Fei Huang}{comp}
\icmlauthor{Yongbin Li}{comp}
% \icmlauthor{Firstname1 Lastname1}{equal,yyy}
% \icmlauthor{Firstname2 Lastname2}{equal,yyy,comp}
% \icmlauthor{Firstname3 Lastname3}{comp}
\end{icmlauthorlist}

\icmlaffiliation{comp}{Alibaba Group}

\icmlcorrespondingauthor{Bowen Yu}{yubowen.ybw@alibaba-inc.com}
\icmlcorrespondingauthor{Yongbin Li}{shuide.lyb@alibaba-inc.com}

\icmlkeywords{Model Merging, Language Models}

\vskip 0.3in
]



\printAffiliationsAndNotice{}  % leave blank if no need to mention equal contribution

\begin{abstract}
In this paper, we unveil that Language Models (LMs) can acquire new capabilities by assimilating parameters from homologous models without retraining or GPUs. We first introduce DARE to set most delta parameters (i.e., the disparity between fine-tuned and pre-trained parameters) to zeros without affecting the abilities of Supervised Fine-Tuning (SFT) LMs, which randomly \textbf{D}rops delta parameters with a ratio $p$ \textbf{A}nd \textbf{RE}scales the remaining ones by $1 / (1 - p)$ to approximate the original embeddings. Then, we use DARE as a versatile plug-in to sparsify delta parameters of multiple SFT homologous models for mitigating parameter interference and merge them into a single model by parameter fusing. We experiment with encoder- and decoder-based LMs, showing that: (1) SFT delta parameter value ranges are typically small (within 0.002) with extreme redundancy, and DARE can effortlessly eliminate 90\% or even 99\% of them; (2) DARE can merge multiple task-specific LMs into one LM with diverse capabilities. Notably, this phenomenon is more pronounced in large-scale LMs, where the merged LM reveals the potential to surpass the performance of any source LM, providing a new discovery. We also utilize DARE to create a merged LM that ranks first among models with 7 billion parameters on the Open LLM Leaderboard.
\end{abstract}

\section{Introduction}
\label{section-1}

\begin{figure}[!htbp]
    \centering
    \includegraphics[width=1.00\columnwidth]{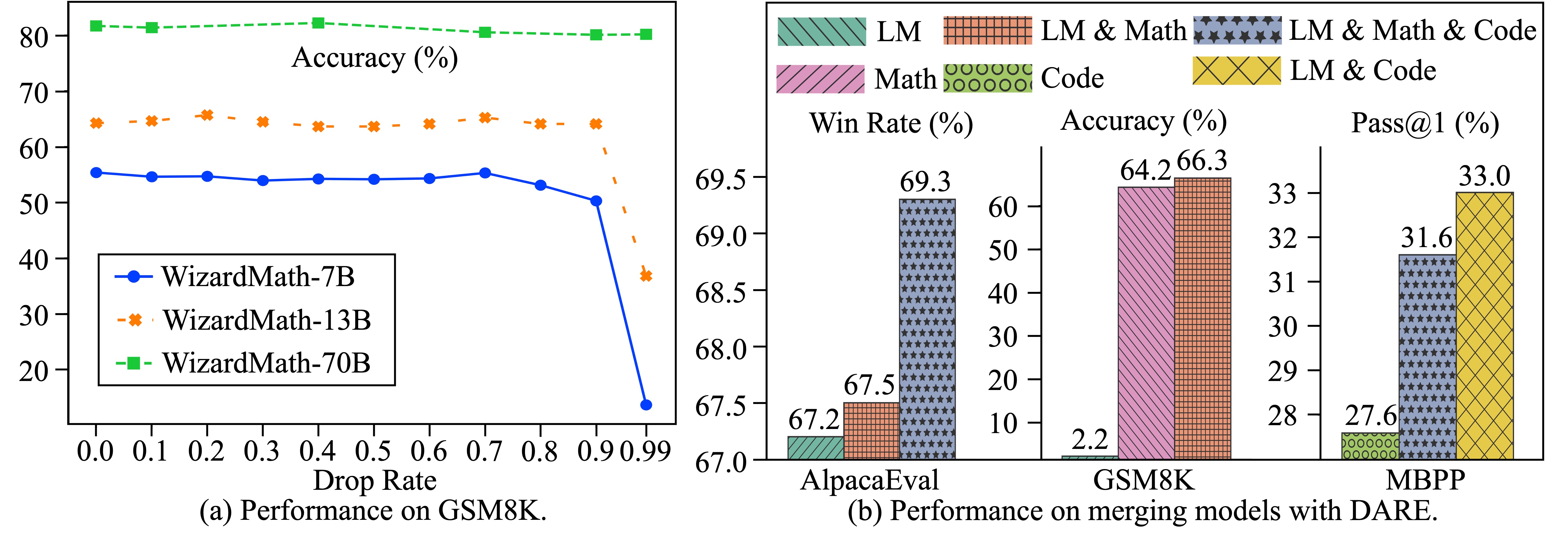}
    \caption{(\textbf{Left}) DARE can effectively eliminate 90\% or even 99\% delta parameters of WizardMath on GSM8K. (\textbf{Right}) DARE can merge multiple task-specific SFT language models into a single model with all the abilities. LM, MATH, and Code are abbreviations of WizardLM-13B, WizardMath-13B, and llama-2-13b-code-alpaca.}
    \label{fig:introduction_llms_merge}
    % \vspace{-5mm}
\end{figure}

Human beings have harbored a longstanding desire to acquire additional abilities through various ways, as expressed in mediums like movies and games. For example, in X-Men's Apocalypse, the character can absorb the powers of other mutants to strengthen himself. Likewise, the protagonist in the Super Mario games can gain superpowers like throwing fireballs by absorbing in-game items. In this paper, we astonishingly find that Language Models (LMs), similar to Apocalypse and Super Mario, can enhance their capabilities by absorbing other models without the need for retraining or even GPUs.

Formally, Supervised Fine-Tuning (SFT) is the most widely adopted strategy for unlocking task-specific abilities to LMs by optimizing their parameters \cite{DBLP:journals/corr/abs-2002-06305,DBLP:journals/corr/abs-2303-18223}. The effectiveness of SFT is fully evident in the alteration of the model parameters before and after SFT, referred to as \textit{delta parameters} \cite{DBLP:journals/natmi/DingQYWYSHCCCYZWLZCLTLS23}. We first show that SFT LM (either encoder- or decoder-based) always tends to acquire excessively redundant delta parameters. To be specific, we present DARE (\textbf{D}rop \textbf{A}nd \textbf{RE}scale), which randomly sets certain delta parameters to zeros with a drop rate $p$ and subsequently rescales the remaining ones by a factor of $1 / (1 - p)$. Although conceptually simple, DARE can eliminate up to 99\% delta parameters with minimal impact on the performance when the LM's parameters reach 70 billion (see \figref{fig:introduction_llms_merge}(a)). Moreover, the more parameters the LM has, the larger $p$ it can tolerate. We attribute the effectiveness of DARE to its ability to approximate the original embeddings, which is verified theoretically and empirically.

Furthermore, we can merge multiple homologous SFT LMs (fine-tuned from the same backbone) based on DARE without compromising their capabilities. As long as a small portion of the delta parameters remain unaffected during merging, the abilities of LMs unlocked by SFT can still be preserved. We first employ DARE to eliminate redundant delta parameters in each model before merging, which can potentially mitigate the interference of parameters among multiple models \cite{DBLP:journals/corr/abs-2306-01708}. Then, we apply established model merging techniques \cite{DBLP:conf/icml/WortsmanIGRLMNF22,DBLP:conf/iclr/IlharcoRWSHF23,DBLP:conf/nips/MatenaR22,DBLP:conf/iclr/Jin0P023,DBLP:journals/corr/abs-2306-01708} to fuse the parameters with reduced redundancy for creating one model with diverse capabilities. 

We conduct extensive experiments with encoder-based LMs on GLUE benchmark, and decoder-based LMs with three distinct abilities: instruction-following, mathematical reasoning, and code-generating. We observe that:

(1) SFT LMs exhibit a substantial number of redundant delta parameters regardless of their backbones (e.g., BERT, RoBERTa, LLaMA, Llama 2, or Code Llama). DARE can \textit{remove 90\% or even 99\% delta parameters} without significantly affecting the model performance. DARE is able to approximate the original embeddings well and provide very similar embeddings for each layer of the LM. The rescale operation is crucial to guarantee the success of DARE, and dropping 30\% or 40\% delta parameters without rescaling would noticeably lead to worse results.

(2) DARE often retains or enhances the performance of various model merging methods on encoder-based LMs. For decoder-based LMs, simply averaging the parameters can already yield satisfactory results. As shown in \figref{fig:introduction_llms_merge}(b), we merge various decoder-based LMs by DARE and Task Arithmetic \cite{DBLP:conf/iclr/IlharcoRWSHF23}, leading to considerable improvements. For example, 3.10\% for LM \& Math \& Code vs. LM on AlpacaEval, 3.18\% for LM \& Math vs. Math on GSM8K, and 19.57\% for LM \& Code vs. Code on MBPP. We also use DARE to create a merged LM with 7 billion parameters, \textit{attaining the top-ranking position on the Open LLM Leaderboard}. It is fascinating that all the benefits are achieved by solely using CPUs without retraining.

(3) SFT delta parameters usually stay within 0.002, indicating minimal modifications to the pre-trained LM, and DARE works for delta parameters with relatively small value ranges. However, once models undergo continuous pre-training, the delta parameters can rapidly reach around 0.03, making DARE infeasible. Moreover, dropping only 10\% fine-tuned parameters (i.e., the combination of pre-trained and delta parameters) would lead to a catastrophic decrease in performance, even approaching zero. This finding further confirms that SFT primarily unlocks the abilities of pre-trained LMs, rather than introducing new capabilities. 

The used resources are publicly available at \url{https://github.com/yule-BUAA/MergeLM}.

\section{Related Work}
\label{section-2}

\textbf{Supervised Fine-tuning of Language Models}. 
SFT of LMs aims to impart pre-trained LMs with particular abilities by optimizing them on task-specific data, which has become the de facto standard paradigm in natural language processing \cite{DBLP:journals/corr/abs-2002-06305,DBLP:journals/corr/abs-2303-18223}. Generally, SFT can be divided into two categories: full fine-tuning \cite{radford2018improving,DBLP:conf/naacl/DevlinCLT19} and parameter-efficient fine-tuning \cite{DBLP:conf/icml/HoulsbyGJMLGAG19,DBLP:journals/corr/abs-2103-10385,DBLP:conf/acl/LiL20,DBLP:conf/emnlp/LesterAC21,DBLP:conf/iclr/HuSWALWWC22}. Indeed, the effects of SFT are reflected by the difference between parameters of LMs before and after SFT, i.e., delta parameters. In this paper, we reveal the extreme redundancy of various SFT LMs' delta parameters by proposing an innovative approach DARE, achieving competitive performance with standard SFT LMs by removing 90\% or even 99\% delta parameters.

\textbf{Network Pruning Technique}. 
With the rapidly increasing size of neural networks, network pruning technique has been widely applied to reduce the computational costs \cite{DBLP:journals/corr/abs-1710-09282,DBLP:journals/ijon/LiangGWSZ21}. The objective of network pruning is to eliminate unnecessary parameters while maintaining the model performance \cite{DBLP:conf/iclr/ZhuG18,DBLP:conf/iclr/LiuSZHD19,DBLP:conf/iclr/FrankleC19,DBLP:journals/corr/abs-1902-09574,DBLP:conf/acl/XiaZC22}. Magnitude-based pruning is one classical pruning method, which selects parameters according to their magnitudes (i.e., absolute parameter values) \cite{han2015learning,DBLP:conf/ijcai/LiQJLT18,DBLP:conf/iclr/LeePMAS21}. To be specific, parameters with magnitudes lower than a certain threshold are removed, and others are preserved. In fact, DARE is relevant to the concept of network pruning as it can also drop parameters. But DARE differs from existing pruning techniques in: (1) DARE focuses on delta parameters while most pruning methods deal with fine-tuned parameters; (2) DARE can work well without any retraining or extra data, which are often inevitably required by pruning methods.

\textbf{Model Merging}. 
Model merging has become a trending research direction in recent years, aiming to merge multiple task-specific models into a single model with diverse abilities \cite{DBLP:conf/icml/WortsmanIGRLMNF22,DBLP:conf/nips/MatenaR22,DBLP:conf/iclr/IlharcoRWSHF23,DBLP:conf/iclr/Jin0P023,DBLP:journals/corr/abs-2306-01708,DBLP:journals/corr/abs-2306-14870}. The superiority of model merging over multi-task learning \cite{DBLP:journals/corr/abs-2009-09796,DBLP:journals/tkde/ZhangY22} (which also intends to obtain one model with several abilities) is that model merging pays attention to the fusion of model parameters without accessing the original training data \cite{DBLP:conf/nips/MatenaR22,DBLP:conf/iclr/Jin0P023}. Average Merging \cite{DBLP:conf/icml/WortsmanIGRLMNF22} is one common model merging approach, which utilizes averaged parameters to construct the merged model. Task Arithmetic \cite{DBLP:conf/iclr/IlharcoRWSHF23} employs a pre-defined scaling term to distinguish the importance of various models. Fisher Merging \cite{DBLP:conf/nips/MatenaR22} performs weighted fusions of parameters, where the weights are calculated by the Fisher information matrix \cite{fisher1922mathematical}. RegMean \cite{DBLP:conf/iclr/Jin0P023} masterly solves model merging by optimizing a linear regression problem with closed-form solutions. TIES-Merging \cite{DBLP:journals/corr/abs-2306-01708} tackles the task conflicts in \citet{DBLP:conf/iclr/IlharcoRWSHF23} by trimming low-magnitude parameters, resolving sign disagreements, and disjointly merging parameters with consistent signs. In this paper, we use DARE as a versatile plug-in for existing model merging methods by first sparsifying delta parameters of several SFT homologous models and then merging them into a single model, which is equipped with the capabilities of all the SFT models.

\section{Methodology}
\label{section-3}

\textbf{SFT Delta Parameters}. Let $\bm{\theta}_{\text{PRE}} \in \mathbb{R}^d$ denote the parameters of a pre-trained LM ($d$ is the parameter dimension), such as LLaMA \cite{DBLP:journals/corr/abs-2302-13971} or Llama 2 \cite{DBLP:journals/corr/abs-2307-09288}. For task $t$, SFT can provide a fine-tuned LM with parameters $\bm{\theta}_{\text{SFT}}^t \in \mathbb{R}^d$ by optimizing the pre-trained model on task-specific data. Give the parameters of both pre-trained LM ($\bm{\theta}_{\text{PRE}}$) and SFT LM ($\bm{\theta}_{\text{SFT}}^t$), delta parameters are defined as the difference between parameters of LMs before and after SFT, i.e., $\bm{\delta}^t=\bm{\theta}_{\text{SFT}}^t - \bm{\theta}_{\text{PRE}} \in \mathbb{R}^d$. Since delta parameters reflect the changes in parameters during the SFT process, analyzing the properties of delta parameters can offer a better understanding of SFT.

\textbf{Model Merging Problem}. Given a set of $K$ tasks $\left\{t_1, t_2, \cdots, t_K\right\}$ and $K$ corresponding SFT models with parameters $\left\{\bm{\theta}_{\text{SFT}}^{t_1}, \bm{\theta}_{\text{SFT}}^{t_2}, \cdots, \bm{\theta}_{\text{SFT}}^{t_K}\right\}$, model merging aims to fuse the parameters of $K$ models into a single model with parameters $\bm{\theta}_{\text{M}}$ that can well handle $K$ tasks simultaneously. Following \citet{DBLP:conf/nips/MatenaR22,DBLP:conf/iclr/Jin0P023,DBLP:journals/corr/abs-2306-01708}, we focus on merging fine-tuned models that are optimized from the same pre-trained backbone.

\subsection{DARE: A Simple Approach for Reducing Delta Parameter Redundancy}\label{section-3-1-dare}
\begin{figure*}[!ht]
    \centering
    \includegraphics[width=1.85\columnwidth]{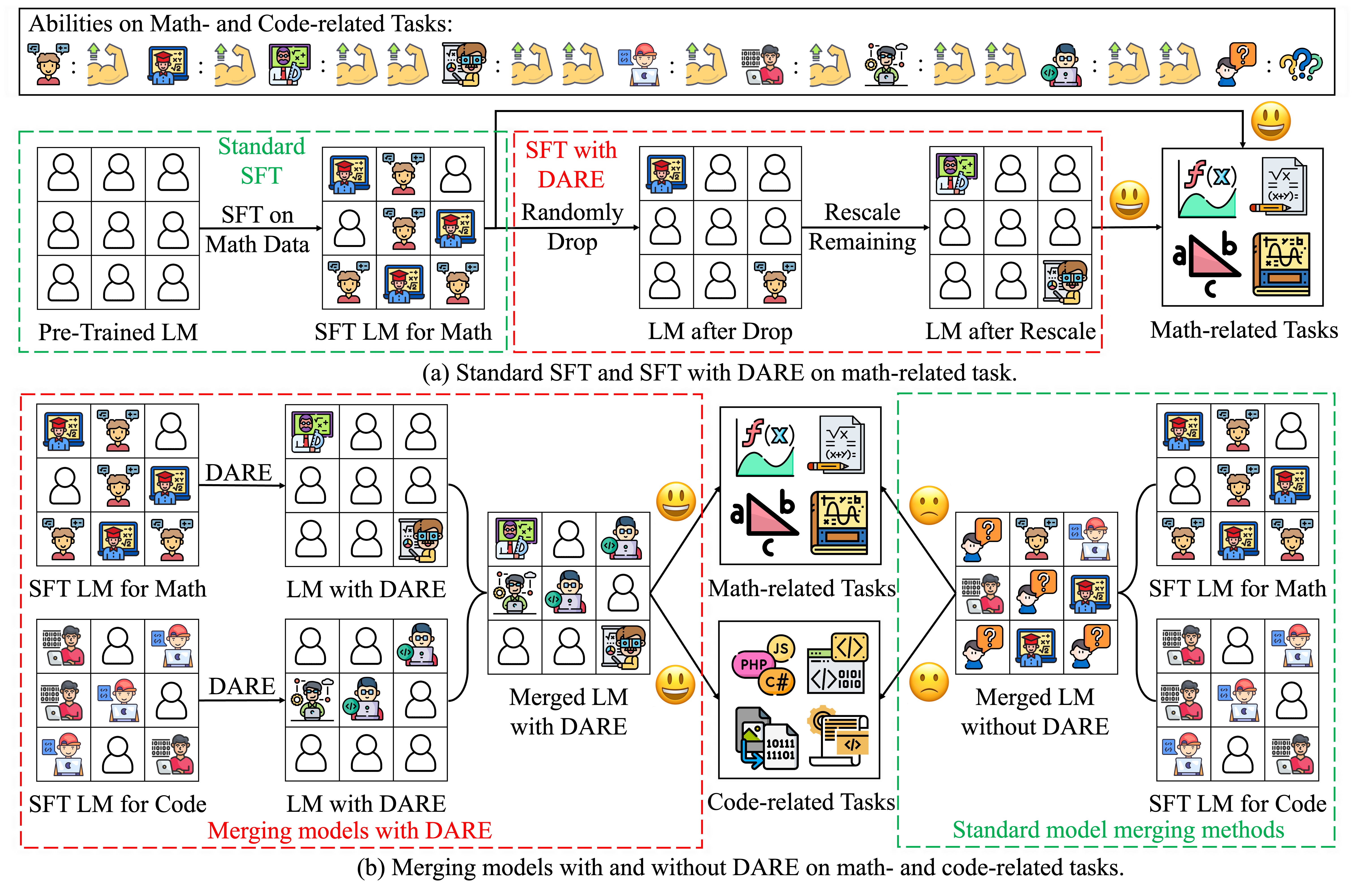}
    \caption{Illustrations of DARE and merging models with DARE. DARE can achieve comparable performance with standard SFT with 90\% or even 99\% delta parameters removed. Moreover, DARE tackles the parameter interference issue when merging models and yields consistent improvements. At the top, we mark each icon with one or two muscle logos, indicating its ability for specific tasks. For example, the first or second icon has one muscle logo for math-related tasks, while the third or fourth icon can perform better in math with two muscle logos. The rescale operation in DARE multiplies the remaining parameters by $1/(1-p)$, which enhances the task-specific abilities and leads to changes in icons after rescaling.}
    \label{fig:framework}
\end{figure*}

In this work, we reveal the extremely redundant properties of the delta parameters of SFT LMs and propose DARE to effectively reduce delta parameter redundancy (see \figref{fig:framework}(a)). DARE is conceptually simple and consists of two steps: drop and rescale. Given delta parameters $\bm{\delta}^t=\bm{\theta}_{\text{SFT}}^t - \bm{\theta}_{\text{PRE}}$, DARE first performs random drop on $\bm{\delta}^t$ based on a drop rate $p$ (setting their values to zeros) and then rescales the remaining ones by a factor of $1 / (1 - p)$ as follows,
\begin{gather} 
\label{equ:drop_rescale}
    \bm{m}^t \sim \text{Bernoulli}(p), \notag \\
    \bm{\widetilde{\delta}}^t = \left(\bm{1} - \bm{m}^t\right) \odot \bm{\delta}^t, \\
    \bm{\hat{\delta}}^t = \bm{\widetilde{\delta}}^t / (1 - p). \notag
\end{gather}
Finally, we combine $\bm{\hat{\delta}}^t$ and $\bm{\theta}_{\text{PRE}}$ via addition to obtain the parameters for inference, i.e., $\bm{\theta_{\text{DARE}}}^t = \bm{\hat{\delta}}^t  + \bm{\theta}_{\text{PRE}}$. We prove that even after removing most delta parameters, DARE can well preserve the model performance by approximating the original embeddings.

\textbf{Theoretical Analysis}. We discuss linear transformation since most parameters of LMs play a role in this basic operation (e.g., the computations in feed-forward networks, the projections of queries, keys, values, and outputs in self-attention modules). Let $\bm{W} / \Delta \bm{W} \in \mathbb{R}^{m \times n}$ and $\bm{b} / \Delta \bm{b} \in \mathbb{R}^{m}$ be the pre-trained/delta parameters. The input is a vector $\bm{x} \in \mathbb{R}^{n}$. Expectation of the $i$-th ($1 \leq i \leq m$) dimension of the original embeddings $\bm{h} \in \mathbb{R}^{m}$ is computed by
\begin{gather*}
\label{equ:original_embedding_expectation}
    \mathbb{E} [h_i] = \mathbb{E} [ \sum \limits_{j=1}^n \left(w_{ij} + \Delta w_{ij}\right) x_j + (b_i + \Delta b_i)] \\
    = \sum \limits_{j=1}^n  x_j \mathbb{E} [w_{ij}] + \mathbb{E} [b_i] + \sum \limits_{j=1}^n x_j \mathbb{E} [\Delta w_{ij}] + \mathbb{E} [\Delta b_i] \\
    = \sum \limits_{j=1}^n w_{ij} x_j + b_i + \sum \limits_{j=1}^n \Delta w_{ij} x_j + \Delta b_i = h_i^{\text{PRE}} + \Delta h_i, \\
\end{gather*}
where $w_{ij} / \Delta w_{ij}$ is the entry located at the intersection of the $i$-th row and $j$-th column within $\bm{W} / \Delta \bm{W}$. Similarly, $b_i / \Delta b_i$ denotes the element positioned at the $i$-th dimension of $\bm{b} / \Delta \bm{b}$. Assuming DARE randomly drops delta parameters with a ratio $p$ and rescales others by a factor of $\gamma$. After using DARE, the delta parameters change to $\Delta \widehat{\bm{W}} \in \mathbb{R}^{m \times n}$ and $\Delta \widehat{\bm{b}} \in \mathbb{R}^{m}$. Therefore, the expectation of the $i$-th dimension of embeddings becomes 
\begin{gather*}
\notag 
\label{equ:rescaled_embedding_expectation}
    \mathbb{E} [\hat{h}_i] = \mathbb{E} [ \sum \limits_{j=1}^n \left(w_{ij} + \Delta \hat{w}_{ij}\right) x_j + (b_i + \Delta \hat{b}_i)] \notag \\
    = \sum \limits_{j=1}^n  x_j \mathbb{E} [w_{ij}] + \mathbb{E} [b_i] + \sum \limits_{j=1}^n x_j \mathbb{E} [\Delta  \hat{w}_{ij}] + \mathbb{E} [\Delta  \hat{b}_i] \notag \\
    = \sum \limits_{j=1}^n  w_{ij} x_j + b_i + \sum \limits_{j=1}^n x_j ((1 - p) \cdot \gamma \cdot \Delta w_{ij} + p \cdot 0) \notag \\ 
    + ((1 - p) \cdot \gamma \cdot \Delta b_i + p \cdot 0) \notag \\ 
    = h_i^{\text{PRE}} + (1 - p) \cdot \gamma \cdot (\sum \limits_{j=1}^n \Delta w_{ij} x_j + \Delta b_i) \notag \\
    = h_i^{\text{PRE}} + (1 - p) \cdot \gamma \cdot \Delta h_i. \notag \\
\end{gather*}
By setting $\gamma = 1 / (1 - p)$, we have $\mathbb{E} [h_i] = \mathbb{E} [\hat{h}_i]$, concluding that DARE can approximate the original embeddings.

\textit{Remark}. We have given a rough proof of why DARE works. In practice, we find that DARE is applicable when the drop rate $p$ is properly set, and the tolerance of $p$ grows with LMs' parameter sizes. Moreover, removing fine-tuned rather than delta parameters would cause a catastrophically decreased performance. A promising future direction is to explore DARE more deeply, such as inferring the upper bound of $p$ with respect to LM capacities and illustrating the intrinsic difference between fine-tuned and delta parameters.

Last, we highlight the connections and differences between DARE and Dropout \cite{DBLP:journals/jmlr/SrivastavaHKSS14}. Both methods involve random dropping and rescaling operations, but they differ in two key aspects: (1) DARE handles delta parameters while Dropout operates on model outputs; (2) DARE aims to reduce delta parameter redundancy \textit{without training}, which \textit{permanently} eliminates delta parameters and only retains others for inference. Dropout is used to prevent models from overfitting, which \textit{temporarily} removes part of outputs during training but preserves all the outputs for inference.

\subsection{Merging Models with DARE}\label{section-3-model-merging-dare}
As DARE effectively reduces the redundancy of delta parameters by setting most of them to zeros, we hypothesize that DARE can help address the interference of parameters when merging multiple models \cite{DBLP:journals/corr/abs-2306-01708}. Take \figref{fig:framework}(b) as an example, when merging math- and code-related models, DARE can assist existing model merging methods to better absorb the abilities of two models with less or no parameter interference.

Formally, given $K$ models that are fine-tuned on $K$ corresponding tasks with parameters $\left\{\bm{\theta}_{\text{SFT}}^{t_1}, \bm{\theta}_{\text{SFT}}^{t_2}, \cdots, \bm{\theta}_{\text{SFT}}^{t_K}\right\}$, we first apply DARE on each parameters $\bm{\theta}_{\text{SFT}}^{t_k}$ ($1 \leq k \leq K$), and derive $\left\{\bm{\theta}_{\text{DARE}}^{t_1}, \bm{\theta}_{\text{DARE}}^{t_2}, \cdots, \bm{\theta}_{\text{DARE}}^{t_K}\right\}$. Then, we adopt established model merging methods to fuse the derived parameters and obtain the merged single model with parameters $\bm{\theta}_{\text{M}}$. Let us take Task Arithmetic \cite{DBLP:conf/iclr/IlharcoRWSHF23} as an instance, whose official computation process is denoted by
\begin{equation}
    \bm{\theta}_{\text{M}} = \bm{\theta}_{\text{PRE}} + \lambda \cdot \sum_{k=1}^K \bm{\delta}^{t_k} = \bm{\theta}_{\text{PRE}} + \lambda \cdot \sum_{k=1}^K (\bm{\theta}_{\text{SFT}}^{t_k} - \bm{\theta}_{\text{PRE}}),
\end{equation}
where $\lambda$ is the scaling term to determine the importance of the models to be merged. When equipped with DARE, the calculation process of Task Arithmetic is rewritten as
\begin{equation}
\begin{split}
   & \bm{\theta}_{\text{DARE}}^{t_k} = \ \text{DARE}\left(\bm{\theta}_{\text{SFT}}^{t_k}, \bm{\theta}_{\text{PRE}}, p\right), \ \text{for} \ 1 \leq k \leq K, \\
   \bm{\theta}_{\text{M}} = \ & \bm{\theta}_{\text{PRE}} + \lambda \cdot \sum_{k=1}^K \bm{\hat{\delta}}^{t_k} = \bm{\theta}_{\text{PRE}} + \lambda \cdot \sum_{k=1}^K (\bm{\theta}_{\text{DARE}}^{t_k} - \bm{\theta}_{\text{PRE}}).
\end{split}
\end{equation}
The expression $\text{DARE}\left(\bm{\theta}_{\text{SFT}}^{t_k}, \bm{\theta}_{\text{PRE}}, p\right)$ signifies the process of deriving delta parameters from $\bm{\theta}_{\text{SFT}}^{t_k}$ and $\bm{\theta}_{\text{PRE}}$, eliminating delta parameters based on drop rate $p$ following \equref{equ:drop_rescale}, and finally combining the sparsified delta parameters with $\bm{\theta}_{\text{PRE}}$ to obtain $\bm{\theta}_{\text{DARE}}^{t_k}$.
In \secref{section-4-merging-methods-with-DARE}, we find that DARE can effectively improve the performance of Task Arithmetic when merging multiple LMs. It is also worth noticing that DARE is a versatile plug-and-play module and can be applied to any model merging methods, such as Average Merging \cite{DBLP:conf/icml/WortsmanIGRLMNF22}, Fisher Merging \cite{DBLP:conf/nips/MatenaR22}, RegMean \cite{DBLP:conf/iclr/Jin0P023}, and TIES-Merging \cite{DBLP:journals/corr/abs-2306-01708}.

\section{Experiments}
\label{section-4}
We conduct extensive experiments on encoder- and decoder-based LMs to show the effectiveness of DARE in reducing SFT delta parameter redundancy and merging models.

\subsection{Experimental Setup}
\textbf{Datasets and Pre-Trained Backbones for Decoder-based LMs}.
We choose AlpacaEval \cite{alpaca_eval} for evaluating instruction-following models (WizardLM \cite{DBLP:journals/corr/abs-2304-12244}). We use GSM8K \cite{DBLP:journals/corr/abs-2110-14168} and MATH \cite{DBLP:conf/nips/HendrycksBKABTS21} for testing mathematical reasoning models (WizardMath \cite{DBLP:journals/corr/abs-2308-09583}). HumanEval \cite{DBLP:journals/corr/abs-2107-03374} and MBPP \cite{DBLP:journals/corr/abs-2108-07732} are adopted for estimating code-generating models (WizardCoder-Python \cite{DBLP:journals/corr/abs-2306-08568} and llama-2-13b-code-alpaca \cite{codealpaca}). These models are fine-tuned based on pre-trained backbones including LLaMA \cite{DBLP:journals/corr/abs-2302-13971}, Llama 2 \cite{DBLP:journals/corr/abs-2307-09288}, and Code Llama \cite{DBLP:journals/corr/abs-2308-12950}. Please see \tabref{tab:llms_SFT_backbone_correspondences} in \secref{section-appendix-llms_SFT_backbone_correspondences} for their versions and correspondences with pre-trained backbones.

\textbf{Datasets and Pre-Trained Backbones for Encoder-based LMs}. For encoder-based LMs, the GLUE benchmark \cite{DBLP:conf/iclr/WangSMHLB19} is used, containing one sentence acceptability dataset CoLA \cite{DBLP:journals/tacl/WarstadtSB19}, one sentiment detection dataset SST-2 \cite{DBLP:conf/emnlp/SocherPWCMNP13}, two paraphrase datasets MRPC \cite{DBLP:conf/acl-iwp/DolanB05} and QQP \cite{shankar2017first}, one sentence similarity dataset STS-B \cite{DBLP:journals/corr/abs-1708-00055}, and three natural language inference datasets MNLI \cite{DBLP:conf/emnlp/BowmanAPM15,DBLP:conf/naacl/WilliamsNB18}, QNLI \cite{DBLP:conf/emnlp/RajpurkarZLL16}, and RTE \cite{DBLP:conf/mlcw/DaganGM05,haim2006second,giampiccolo2007third,bentivogli2009fifth}. 
As the test labels of GLUE are not publicly available, we split the original training data into training and validation sets with ratios of 90\% and 10\%. The original validation data is used as the test set. We choose bert-base-uncased \cite{DBLP:conf/naacl/DevlinCLT19} and roberta-base \cite{DBLP:journals/corr/abs-1907-11692} as pre-trained backbones, and further fine-tune them to get SFT models on the eight datasets.

\textbf{Evaluation Metrics}. 
We calculate win rate for AlpacaEval, zero-shot accuracy for GSM8K and MATH, pass@1 for HumanEval and MBPP, Matthews correlation coefficient for CoLA, accuracy for SST-2, QNLI, and RTE, matched accuracy for MNLI, accuracy and F1 score for MRPC and QQP, and Pearson and Spearman correlation for STS-B.

\textbf{Implementation Details}. 
Following \citet{DBLP:journals/corr/abs-2304-12244,DBLP:journals/corr/abs-2308-09583,DBLP:journals/corr/abs-2306-08568}, the inference of decoder-based LMs is implemented by vLLM \cite{DBLP:conf/sosp/KwonLZ0ZY0ZS23}.
Temperature is set to 0.0 for greedy decoding. The maximal number of generated tokens is 1,024 on GSM8K, and 2,048 on the other four datasets. 
For encoder-based LMs, We fine-tune bert-base-uncased and roberta-base for 10 epochs with a warmup strategy.
The weight decay is 0.01. We use 1e-5 and 5e-5 as learning rates and list the optimal setting of each fine-tuned model in \tabref{tab:plms_learning_rate_configuration} in \secref{section-appendix-plms_learning_rate_configuration}. 
Experiments are conducted on NVIDIA Tesla V100 and A100 GPUs.

\subsection{Extreme Redundancy in SFT Delta Parameters}
\begin{figure}[!htbp]
    \centering
    \includegraphics[width=1.00\columnwidth]{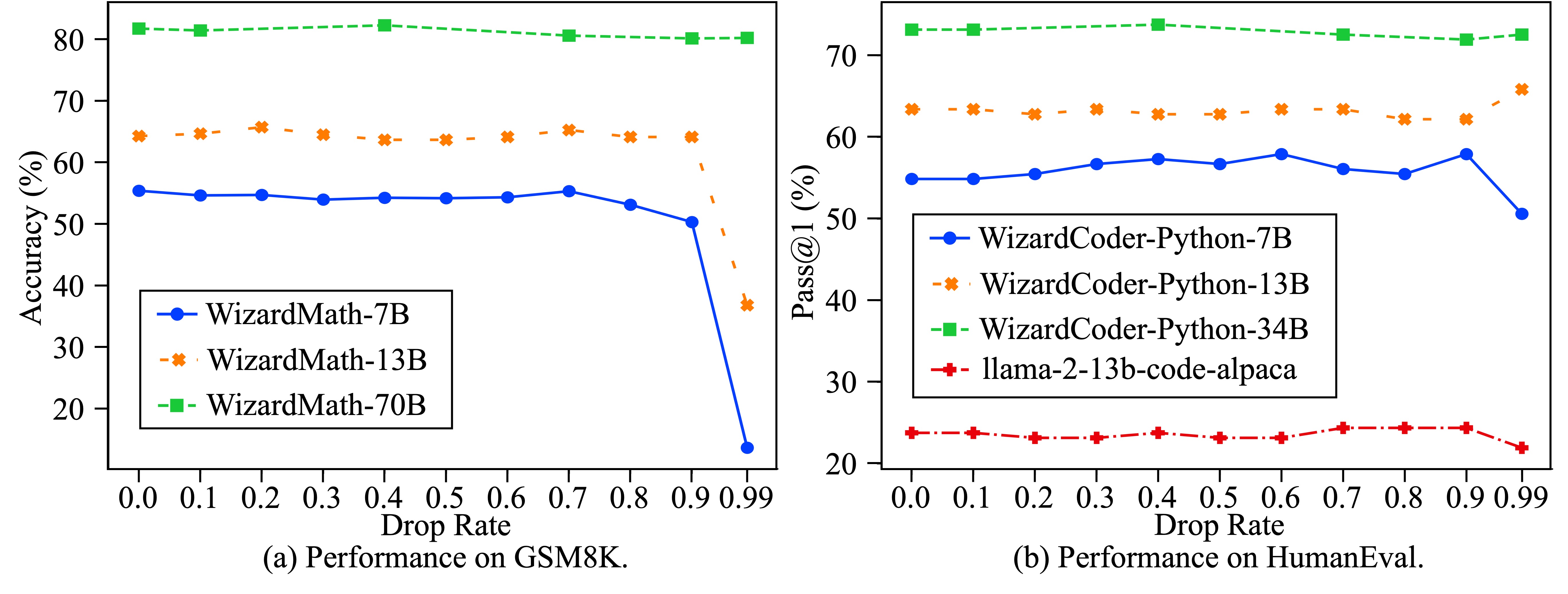}
    \caption{Performance of decoder-based LMs on GSM8K and HumanEval with various drop rates.}
    \label{fig:llms_drop_rate_scaling_curve}
\end{figure}

\begin{figure}[!htbp]
    \centering
    \includegraphics[width=1.0\columnwidth]{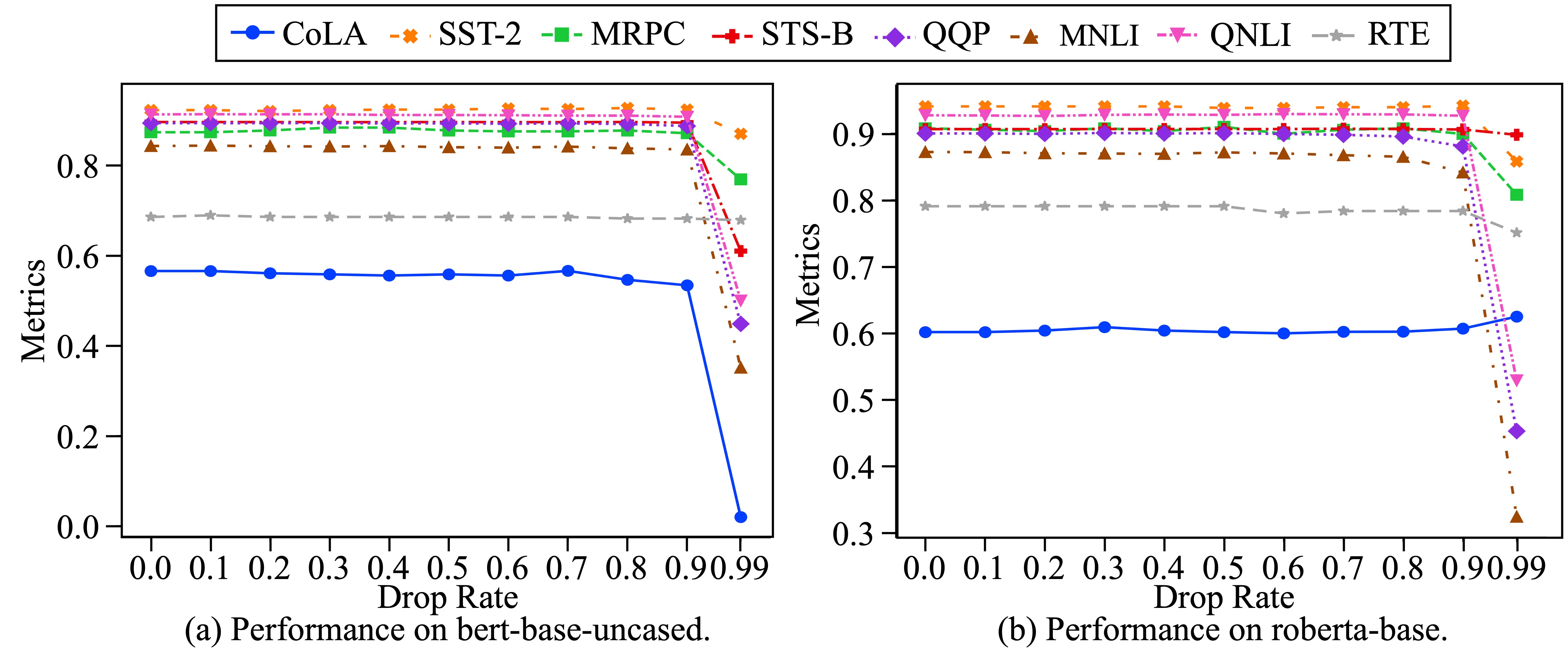}
    \caption{Performance of encoder-based LMs on GLUE with different drop rates.}
    \label{fig:plms_drop_rate_curve}
\end{figure}

We show the extremely redundant property of SFT delta parameters of both decoder- and encoder-based LMs. We vary drop rate $p$ in [0.0, 0.1, 0.2, $\cdots$, 0.9, 0.99] and apply DARE to get models after removing the corresponding ratio of delta parameters. When $p$ is equal to 0.0, we actually obtain the standard SFT LMs. We report the performance of decoder-based LMs on GSM8K and HumanEval as well as encoder-based LMs on eight GLUE datasets in \figref{fig:llms_drop_rate_scaling_curve} and \figref{fig:plms_drop_rate_curve}. Please see results of decoder-based LMs on AlpacaEval, MATH, and MBPP in \figref{fig:llms_drop_rate_scaling_curve_AlpacaEval_MATH_MBPP} in \secref{section-appendix-additional_results_llms_drop_rate_scaling_curve_AlpacaEval_MATH_MBPP}.

\begin{table*}[!htbp]
\centering
\caption{Performance of merging decoder-based WizardLM-13B (LM), WizardMath-13B (Math), and llama-2-13b-code-alpaca (Code) on all the datasets. The best and second-best results are marked in \textbf{bold} and \underline{underlined} fonts.}
\label{tab:llms_merging_all_results}
\resizebox{0.75\textwidth}{!}
{
% \setlength{\tabcolsep}{1.5mm}
% {
\begin{tabular}{c|c|c|c|cc|cc}
\hline
\multirow{2}{*}{\begin{tabular}[c]{@{}c@{}}Merging \\ Methods\end{tabular}} & \multirow{2}{*}{Models} & \multirow{2}{*}{\begin{tabular}[c]{@{}c@{}} Use \\ DARE \end{tabular}} & \begin{tabular}[c]{@{}c@{}}Instruction-\\ following\end{tabular} & \multicolumn{2}{c|}{\begin{tabular}[c]{@{}c@{}}Mathematical \\ Reasoning\end{tabular}} & \multicolumn{2}{c}{\begin{tabular}[c]{@{}c@{}}Code-generating\end{tabular}} \\ \cline{4-8} 
                                                                            &                         &                                                                                  & AlpacaEval                                                       & GSM8K                                          & MATH                                  & HumanEval                              & MBPP                                  \\ \hline
\multirow{3}{*}{\begin{tabular}[c]{@{}c@{}}/ \\ \end{tabular}}    & LM                      & No                                                                                & 67.20                                                            & 2.20                                           & 0.04                                  & 36.59                                  & 34.00                                 \\ \cline{2-8} 
                                                                            & Math                    & No                                                                                & /                                                                & 64.22                                          & 14.02                                 & /                                      & /                                     \\ \cline{2-8} 
                                                                            & Code                    & No                                                                                & /                                                                & /                                              & /                                     & 23.78                                  & 27.60                                 \\ \hline
\multirow{8}{*}{\begin{tabular}[c]{@{}c@{}}Task \\ Arithmetic\end{tabular}} & LM               & No                                                                               & 67.04                                                            & \textbf{66.34}                                 & \underline{13.40}                                 & \underline{28.66}                                  & 30.60                                 \\
                                                                            &          \& Math               & Yes                                                                              & \textbf{67.45}                                                   & \underline{66.26}                                          & 12.86                                 & 26.83                                  & \underline{32.40}                                 \\ \cline{2-8} 
                                                                            & LM              & No                                                                               & \textbf{68.07}                                                   & /                                              & /                                     & \underline{31.70}                                  & \underline{32.40}                                 \\
                                                                            &          \& Code               & Yes                                                                              & \underline{67.83}                                                            & /                                              & /                                     & \textbf{35.98}                         & \textbf{33.00}                        \\ \cline{2-8} 
                                                                            & Math            & No                                                                               & /                                                                & \underline{64.67}                                          & \underline{13.98}                                 & 8.54                                   & 8.60                                  \\
                                                                            &         \& Code                & Yes                                                                              & /                                                                & \textbf{65.05}                                 & 13.96                                 & \underline{10.37}                                  & \underline{9.80}                                  \\ \cline{2-8} 
                                                                            & LM \& Math        & No                                                                               & \underline{69.03}                                                            & \underline{58.45}                                          & 9.88                                  & 18.29                                  & \underline{29.80}                                 \\
                                                                            &      \& Code                   & Yes                                                                              & \textbf{69.28}                                                   & 56.48                                          & \underline{10.16}                                 & \underline{23.17}                                  & \textbf{31.60}                        \\ \hline
\multirow{8}{*}{\begin{tabular}[c]{@{}c@{}}TIES- \\ Merging\end{tabular}} & LM               & No                                                                               & \underline{68.63}                                                            & 15.77                                 & 2.04                                 & \textbf{37.80}                                  & \underline{35.60}                                 \\
                                                                            &          \& Math               & Yes                                                                              & \textbf{68.70}                                                   & \underline{36.16}                                          & \underline{4.56}                                 & 36.59                                  & \textbf{37.00}                                 \\ \cline{2-8} 
                                                                            & LM              & No                                                                               & 63.63                                                   & /                                              & /                                     & 0.0                                  & 0.0                                 \\
                                                                            &          \& Code               & Yes                                                                              & \underline{67.15}                                                            & /                                              & /                                     & \underline{18.29}                         & \underline{26.40}                        \\ \cline{2-8} 
                                                                            & Math            & No                                                                               & /                                                                & 63.23                                          & 13.56                                 & 9.76                                   & 22.40                                  \\
                                                                            &         \& Code                & Yes                                                                              & /                                                                & \textbf{64.82}                                 & \underline{13.88}                                 & \underline{10.37}                                  & \underline{23.60}                                  \\ \cline{2-8} 
                                                                            & LM \& Math        & No                                                                               & 65.91                                                            & \underline{62.55}                                          & \underline{9.54}                                  & 21.95                                  & \underline{30.40}                                 \\
                                                                            &      \& Code                   & Yes                                                                              & \textbf{72.50}                                                   & 58.00                                          & 9.20                                 & \textbf{29.27}                                  & \textbf{31.40}                        \\ \hline
\end{tabular}
}
% }
\end{table*}

We conclude that: (1) \textit{the SFT delta parameters of both encoder- and decoder-based LMs are highly redundant}. DARE can effectively remove 90\% delta parameters without significantly decreasing the performance. In some cases, the drop rate $p$ can even reach 99\%;
(2) \textit{the tolerance of drop rate increases with the sizes of LMs, i.e., LMs with more parameters can withstand higher drop rate}. For example, WizardMath-70B performs well when $p = 0.99$ while WizardMath-7B and WizardMath-13B fail. This depicts some connections with the scaling laws of LMs \cite{DBLP:journals/corr/abs-2001-08361,DBLP:journals/corr/abs-2203-15556}, indicating that there may exist quantifiable correlations between model sizes and drop rates they can afford. 

\subsection{Merging Models with DARE on SFT LMs}\label{section-4-merging-methods-with-DARE}
\begin{figure}[!htbp]
    \centering
    \includegraphics[width=1.00\columnwidth]{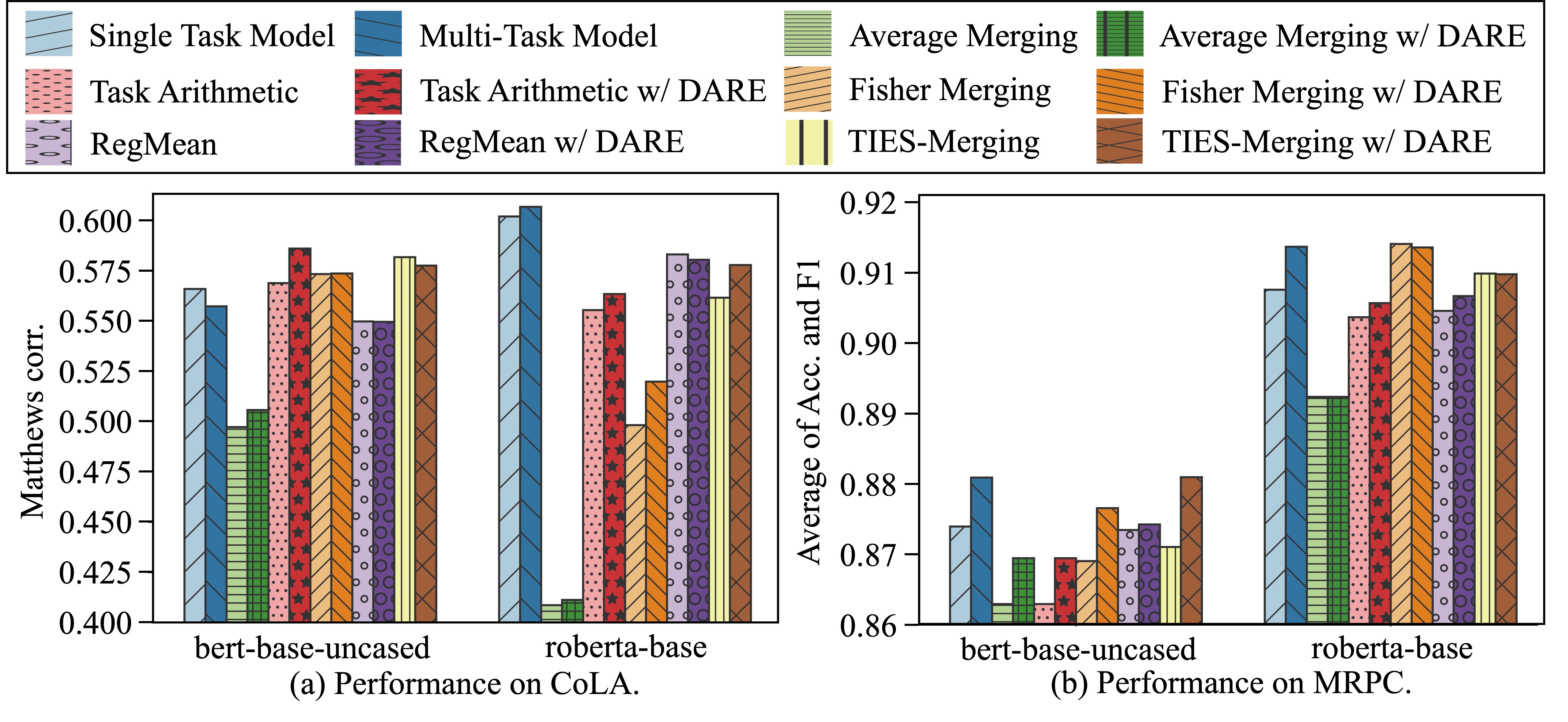}
    \caption{Performance of merging encoder-based bert-base-uncased and roberta-base on CoLA and MRPC.}
    \label{fig:plms_model_merging}
    % \vspace{-5mm}
\end{figure}

We combine DARE with five model merging methods, including Average Merging \cite{DBLP:conf/icml/WortsmanIGRLMNF22}, Task Arithmetic \cite{DBLP:conf/iclr/IlharcoRWSHF23}, Fisher Merging \cite{DBLP:conf/nips/MatenaR22}, RegMean \cite{DBLP:conf/iclr/Jin0P023}, and TIES-Merging \cite{DBLP:journals/corr/abs-2306-01708}. Please see \secref{section-appendix-model_merging_methods_descriptions} for more descriptions of the methods. For feasible computations, we merge decoder-based LMs based on Task Arithmetic and TIES-Merging. The scaling term in both methods is chosen from [0.5, 1.0], and the retain ratio of largest-magnitude parameters in TIES-Merging is selected from [0.5, 0.7, 0.9]. We merge WizardLM-13B, WizardMath-13B, and llama-2-13b-code-alpaca since all of them adopt Llama-2-13b as the pre-trained backbone. WizardCoder-Python-13B is not selected as it is fine-tuned from CodeLlama-13b-Python. We merge encoder-based LMs with all five methods and perform grid search on some hyperparameters (see \tabref{tab:plms_hyperparameter_searched_ranges_merging_methods} in \secref{section-appendix-plms_grid_search_model_merging_hyperparameters} for more details). Following \citet{DBLP:conf/iclr/Jin0P023,DBLP:journals/corr/abs-2306-01708}, we also fine-tune the models under the multi-task learning setting and report the oracle results. We show the performance of merging decoder-based LMs in \tabref{tab:llms_merging_all_results} and present partial results of merging encoder-based LMs in \figref{fig:plms_model_merging}. Please refer to \figref{fig:plms_model_merging_all_results} in \secref{section-appendix-additional_results_lms_model_merging_DARE} for the complete results.

From \tabref{tab:llms_merging_all_results}, we find that: 1) DARE often facilitates Task Arithmetic and TIES-Merging on merging decoder-based LMs, which even yields better results than the source model in many cases, offering a novel discovery unobserved in previous works. For instance, the improvements brought by Task Arithmetic with DARE are 3.10\% for LM \& Math \& Code vs. LM on AlpacaEval, 3.18\% for LM \& Math vs. Math on GSM8K, and 19.57\% for LM \& Code vs. Code on MBPP; 2) Compared with Task Arithmetic, TIES-Merging tends to benefit more from DARE. This is because TIES-Merging first eliminates delta parameters with lower magnitudes for each model, which potentially decreases the performance. When using DARE, delta parameters can be effectively removed by resetting them to zeros without adversely affecting the performance. Thus, TIES-Merging just drops delta parameters sparsified by DARE (with zero as the smallest magnitude), avoiding performance reduction in the first step; 3) It seems that llama-2-13b-code-alpaca is not well fine-tuned for generating codes since it performs worse than WizardLM-13B, which may affect the model merging performance. We additionally evaluate the code-generating ability of the merger of WizardLM-13B and WizardMath-13B, which obtains better results than llama-2-13b-code-alpaca, explaining the suboptimal performance of the amalgamation of WizardMath-13B and llama-2-13b-code-alpaca. Therefore, an essential prerequisite for effective model merging is that each source model to be merged should be well fine-tuned.

From \figref{fig:plms_model_merging}, we observe that DARE often yields modestly better results of various merging methods, achieving an average improvement of 0.58\%, 0.36\%, 0.37\%, -0.03\%, and 0.84\% on Average Merging, Task Arithmetic, Fisher Merging, RegMean, and TIES-Merging. However, the merged model still struggles to surpass the single model in some cases, which is in line with the conclusions in \citet{DBLP:conf/nips/MatenaR22,DBLP:conf/iclr/Jin0P023,DBLP:journals/corr/abs-2306-01708}.

Last but not least, from both \tabref{tab:llms_merging_all_results} and \figref{fig:plms_model_merging}, we further conclude that the improvements caused by DARE are more pronounced in decoder-based LMs compared to encoder-based LMs. One possible reason is that decoder-based LMs are able to accommodate more abilities than encoder-based LMs due to their substantially larger sizes.

We further verify the effectiveness of DARE in merging decoder-based LMs apart from the Llama 2 backbone (e.g., Mistral-7B \cite{DBLP:journals/corr/abs-2310-06825}). We provide two merged decoder-based LMs with 7 billion parameters (namely, supermario\_v1 and supermario\_v2) and evaluate them on Open LLM Leaderboard \cite{open-llm-leaderboard}. Please see \secref{section-appendix-details_7b_merged_model_leaderboard} for more details of the source models and benchmarks. 
\begin{table}[!htbp]
\centering
\caption{Results of 7B LMs on the Open LLM Leaderboard.}
\label{tab:llms_merging_open_llm_leaderboard_results}
\resizebox{1.00\columnwidth}{!}
{
\setlength{\tabcolsep}{0.6mm}
{
\begin{tabular}{c|ccccccc}
\hline
Models                & Average & ARC   & Hella. & MMLU  & TQA & Wino. & GSM8K \\ \hline
NeuralBeagle14-7B     & 74.74   & 72.95 & 88.34     & 64.55 & 69.93      & 82.40       & 70.28 \\
Beagle14-7B        & 74.76   & 72.95 & 87.95     & 64.70 & 68.88      & 82.64      & 71.42 \\
supermario\_v1        & \textbf{74.85}   & 73.72 & 88.71     & 64.57 & 68.23      & 85.64      & 68.23 \\ \hline
WildMarcoroni-7B     & 75.29   & 73.98 & 88.61     & 64.81 & 69.76      & 84.29       & 70.28 \\
WestSeverus-7B & 75.29   & 71.42 & 88.27     & 64.79 & 72.37      & 83.27      & 71.65 \\
supermario\_v2        & \textbf{75.49}   & 72.95 & 88.53     & 64.99 & 71.22      & 83.90      & 71.34 \\ \hline
\end{tabular}
}
}
\end{table}
From \tabref{tab:llms_merging_open_llm_leaderboard_results}, we find that the merged LMs beat the source models they are built upon, achieving a certain degree of improvement. Notably, until January 28th, 2024, \textit{supermario\_v2 achieves the first rank on the Open LLM Leaderboard}. It is thrilling that these benefits can be cheaply acquired by merely utilizing CPUs.

\subsection{Importance of the Rescale Operation}\label{section-4-rescale-operation}
\begin{figure}[!htbp]
    \centering
    \includegraphics[width=1.00\columnwidth]{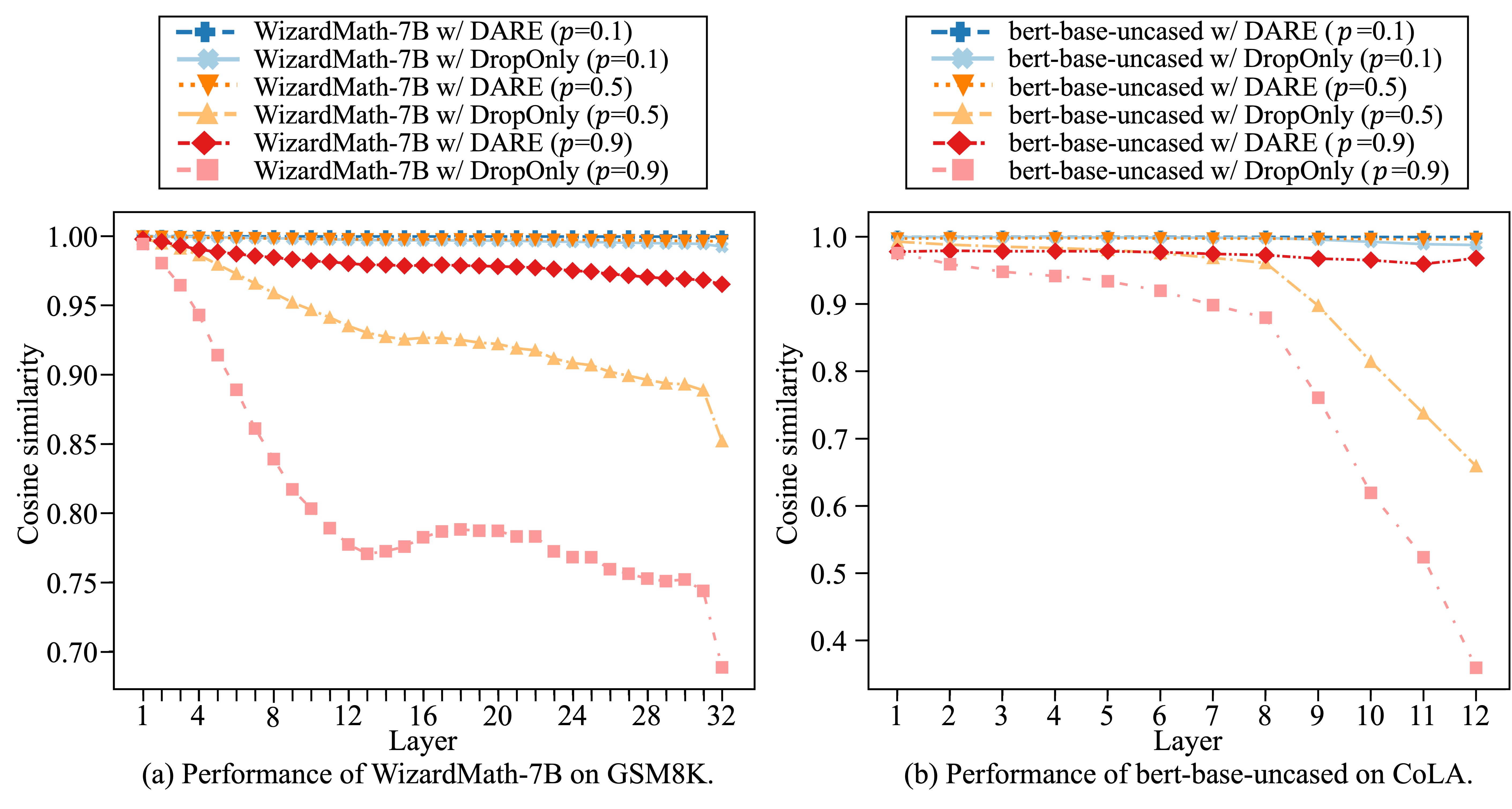}
    \caption{Cosine similarities of each layer's embeddings between the original LM and LM with DARE or DropOnly.}
    \label{fig:lms_all_layers_embedding_cosine_similarities}
\end{figure}

As analyzed in \secref{section-3-1-dare}, the rescale operation in DARE is essential to approximate the original embeddings. To verify this, we introduce DropOnly which randomly drops delta parameters without rescaling. We calculate the similarities of embeddings between the original LM and LM with DARE or DropOnly. Specifically, we obtain the embeddings of each input token layer-by-layer and report the average cosine similarities. Results of WizardMath-7B on GSM8K and bert-base-uncased on CoLA are shown in \figref{fig:lms_all_layers_embedding_cosine_similarities}. 

\begin{figure}[!htbp]
    \centering
    \includegraphics[width=1.00\columnwidth]{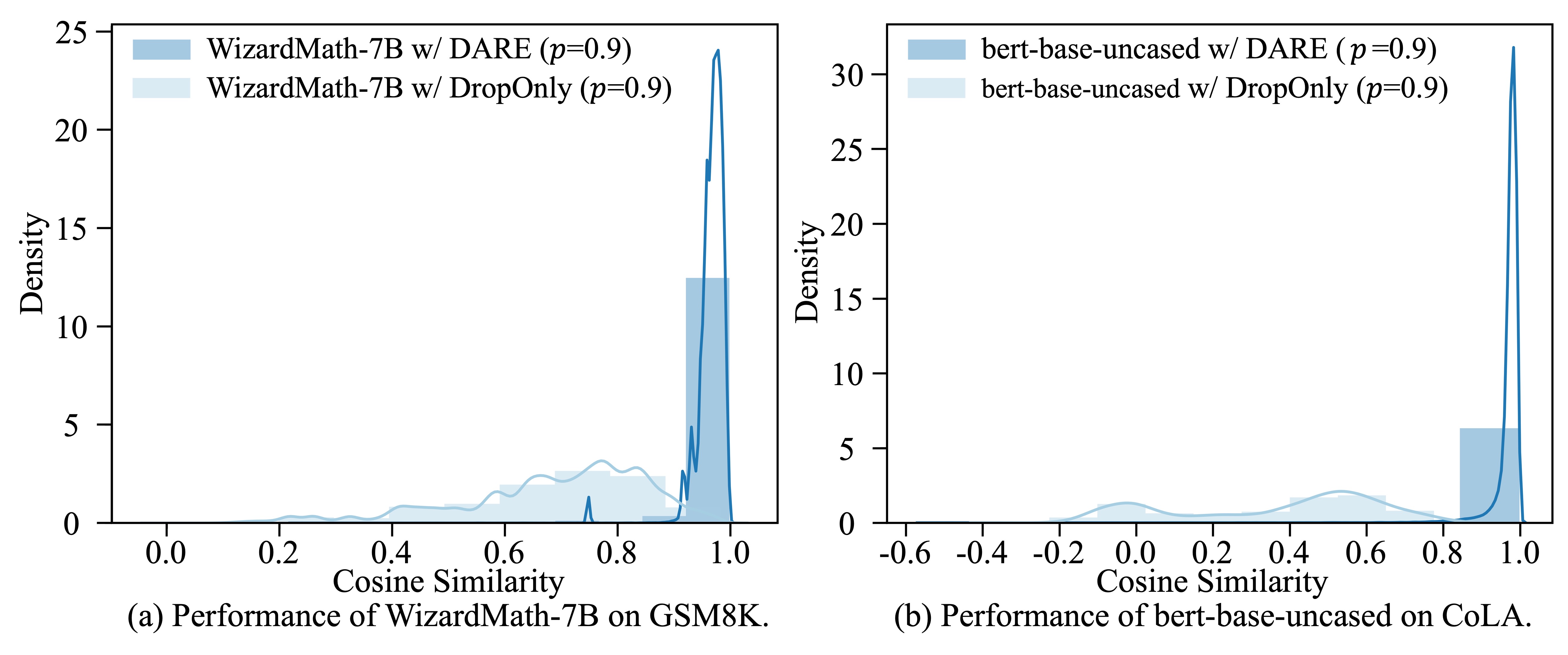}
    \caption{Distributions of cosine similarities of the last layer's embeddings between the original LM and LM with DARE or DropOnly.}
    \label{fig:lms_last_layer_embedding_cosine_similarities_distributions}
\end{figure}

We observe that DARE can perfectly maintain the original embeddings in each layer with similarities higher than 0.95 even when removing 90\% delta parameters. However, DropOnly just preserves the original embeddings with $p=0.1$ and the similarities sharply decline when $p$ is higher. For example, the similarities on WizardMath-7B decrease to about 0.85/0.68 when $p$ is 0.5/0.9).
We further show the distributions of embeddings' cosine similarities in the last layer in \figref{fig:lms_last_layer_embedding_cosine_similarities_distributions}, demonstrating the ability of DARE in approximating original embeddings. Note that similar findings can be obtained on other LMs and datasets but they are not presented due to page limits. 

\begin{figure}[!htbp]
    \centering
    \includegraphics[width=1.00\columnwidth]{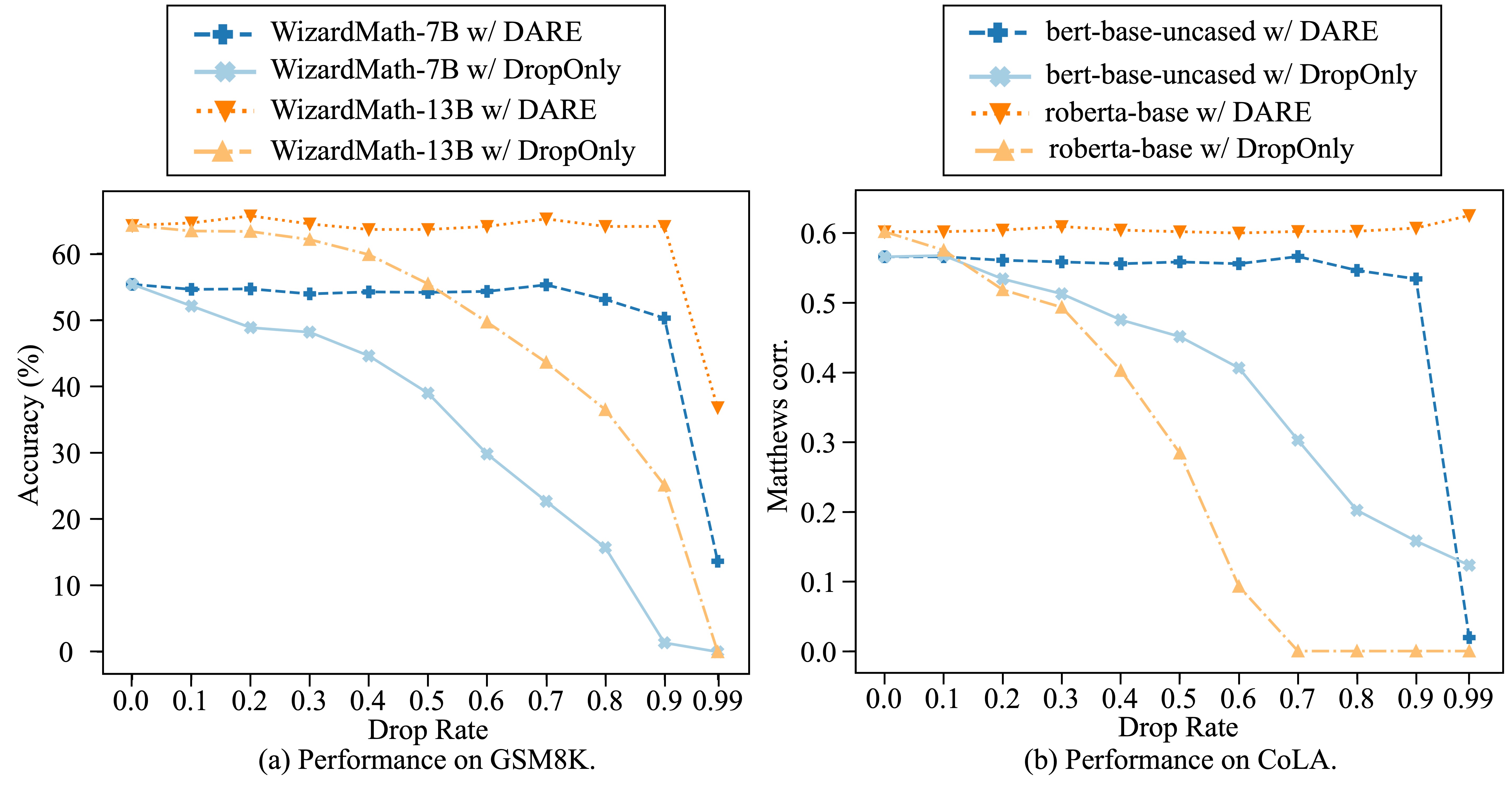}
    \caption{Comparisons between DARE and DropOnly on GSM8K and CoLA on various LMs.}
    \label{fig:lms_rescale_comparison}
\end{figure}

We also report the performance of LMs with DARE and DropOnly in \figref{fig:lms_rescale_comparison}. See \figref{fig:llms_rescale_comparison_AlpacaEval_MATH_HumanEval_MBPP} and \figref{fig:plms_rescale_comparison_all_results} in \secref{section-appendix-additional_results_comparisons_DARE_with_DropOnly} for additional results. We observe that discarding the rescale operation usually leads to worse results, and the performance gaps between DARE and DropOnly become more significant with the increase of $p$. This validates the effectiveness of the rescale operation in DARE once again.

\subsection{Comparison with Magnitude-based Pruning}
\begin{figure}[!htbp]
    \centering
    \includegraphics[width=1.00\columnwidth]{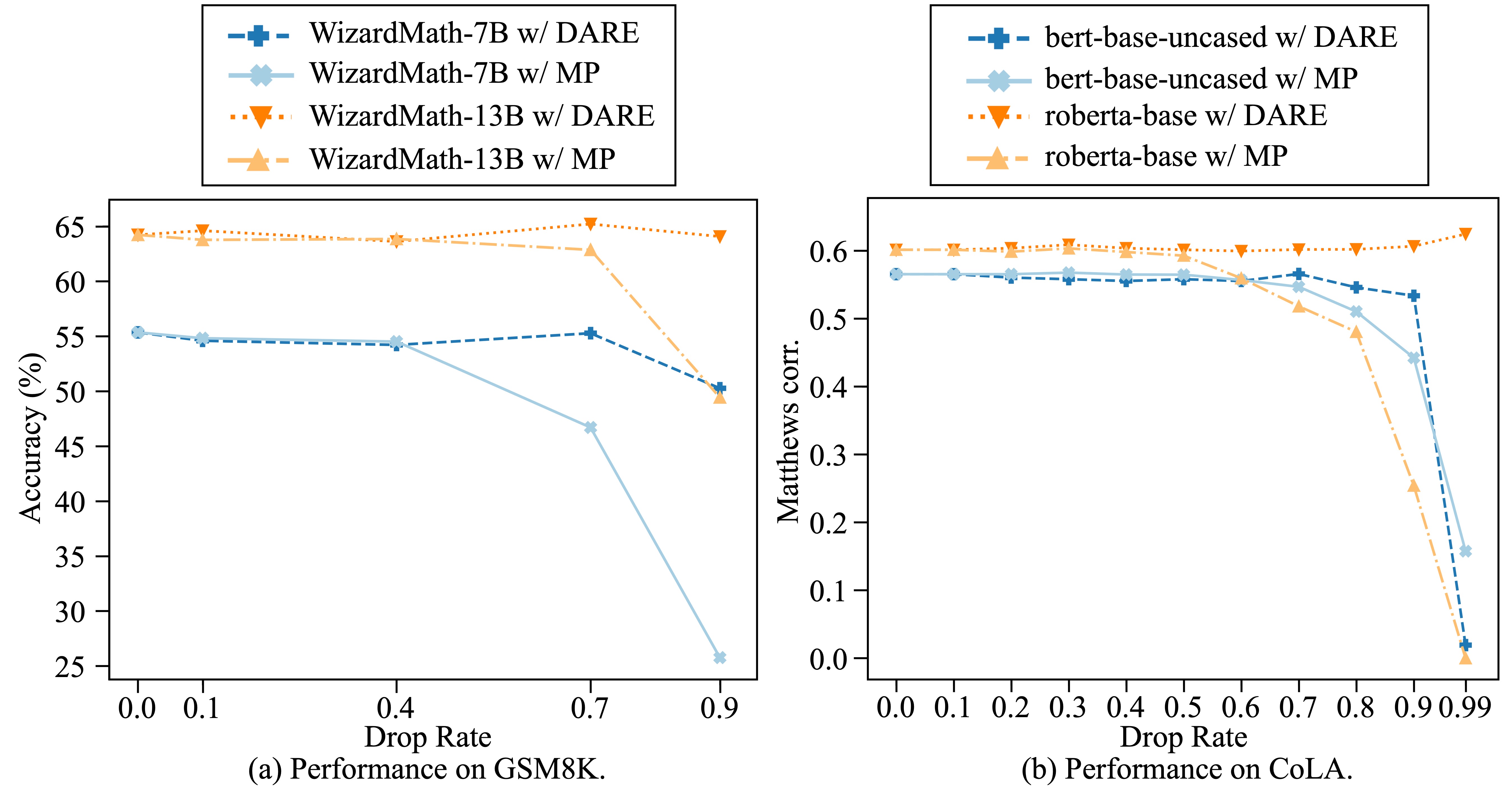}
    \caption{Comparisons between DARE and MP on GSM8K and CoLA on various LMs.}
    \label{fig:lms_magnitude_pruning_comparison}
\end{figure}

We compare DARE with the commonly used Magnitude-based Pruning (MP) \cite{han2015learning,DBLP:conf/ijcai/LiQJLT18,DBLP:conf/iclr/LeePMAS21}, which chooses parameters based on their magnitudes. For more fair and credible comparisons, we adapt MP to operate on delta parameters and discard the retraining process. We show partial results of LMs with DARE and MP in \figref{fig:lms_magnitude_pruning_comparison}. Please refer to \figref{fig:llms_magnitude_pruning_comparison_AlpacaEval_MATH_HumanEval_MBPP} and \figref{fig:plms_magnitude_pruning_comparison_all_results} in \secref{section-appendix-additional_results_comparisons_DARE_with_MP} for extra results. 

We find that DARE outperforms MP in most cases and the superiority of DARE is more obvious when the drop rate becomes higher, verifying the superiority of DARE in abandoning delta parameters. The reason is that MP fails to preserve the original embeddings since it neglects the contributions of delta parameters with lower magnitudes. We have also tried to combine MP with the rescale operation but got worse results than using MP separately. For example, when the drop rate is 0.7, the performance of MP on 7B LMs decreases from 43.85 to 10.61 on AlpacaEval, from 46.70 to 0.37 on GSM8K, and from 21.34 to 3.05 on HumanEval. This is because MP removes parameters with smaller magnitudes and retains certain parameters with larger magnitudes. Simply rescaling the remaining parameters would result in unpredictable performance.

\subsection{When Can DARE Be Used?}
\begin{figure}[!htbp]
    \centering
    \includegraphics[width=0.95\columnwidth]{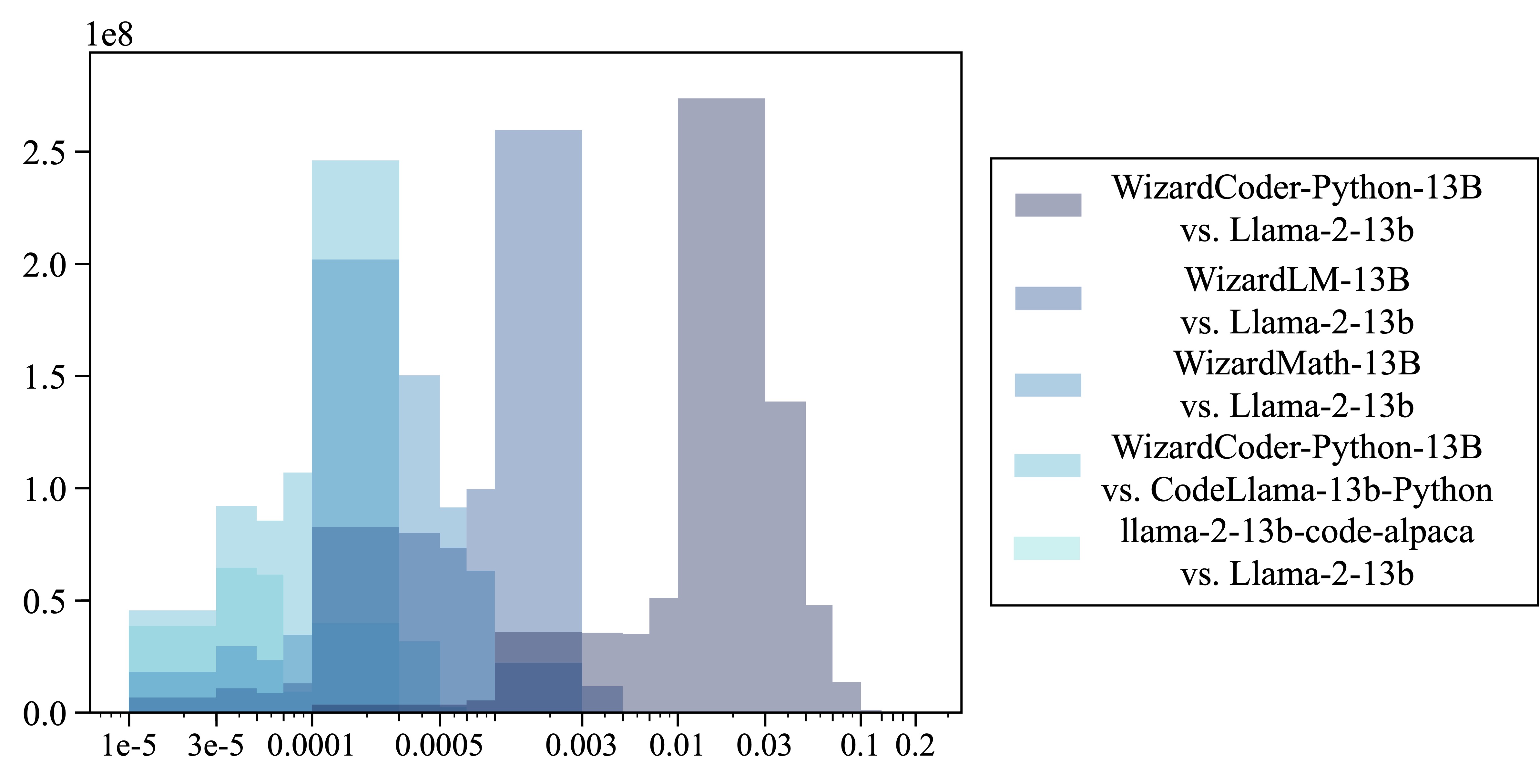}
    \caption{Delta parameter absolute values of 13B decoder-based LMs vs. the pre-trained backbones.}
    \label{fig:llms_absolute_weight_range}
\end{figure}

We investigate the prerequisites that DARE can work. We choose Llama-2-13b instead of CodeLlama-13b-Python as the pre-trained backbone for WizardCoder-Python-13B and apply DARE to derive the model after dropping certain delta parameters for evaluation. We find that the pass@1 metric on HumanEval/MBPP drastically decreases from 63.41/55.4 to 0.0/0.0 when only 10\% delta parameters are removed. We deduce this is because Code Llama models are additionally trained with 500B tokens of code-related data \cite{DBLP:journals/corr/abs-2308-12950}, resulting in more obvious changes in parameter values with respect to Llama 2 models. Since WizardCoder-Python-13B is fine-tuned based on CodeLlama-13b-Python, when it uses Llama-2-13b as the pre-trained backbone, the ranges of SFT delta parameters would become much larger, making DARE infeasible. To verify this, we depict the \textit{absolute values} of SFT delta parameters of 13B decoder-based LMs vs. various pre-trained backbones in \figref{fig:llms_absolute_weight_range}. Please see \figref{fig:llms_parameter_change_range}, \figref{fig:bert_parameter_change_range} and \figref{fig:roberta_parameter_change_range} in \secref{section-appendix-additional_results_original_delta_parameter_ranges} for the SFT delta parameter ranges on decoder- and encoder-based LMs. Additionally, we present the statistics on the percentiles of delta parameter ranges of both decoder- and encoder-based LMs in \tabref{tab:statistics_parameters_changed_ranges} in \secref{section-appendix-additional_results_original_delta_parameter_ranges}.

From the results, we observe the absolute values of delta parameters of WizardCoder-Python-13B vs. Llama-2-13b (often greater than 0.01) are several orders of magnitude bigger than those of WizardCoder-Python-13B vs. CodeLlama-13b-Python (usually within 0.0002), causing the failure of DARE. For other 13B decoder-based LMs fine-tuned from Llama-2-13b, most of their absolute values of delta parameters are less than 0.002, making DARE a proper choice. To this end, we conclude that DARE can work well when the absolute values of SFT delta parameters are relatively small (e.g., within 0.002). Otherwise, DARE may fail. 

\subsection{Can DARE Drop Fine-tuned Parameters?}
\begin{figure}[!htbp]
    \centering
    \includegraphics[width=1.00\columnwidth]{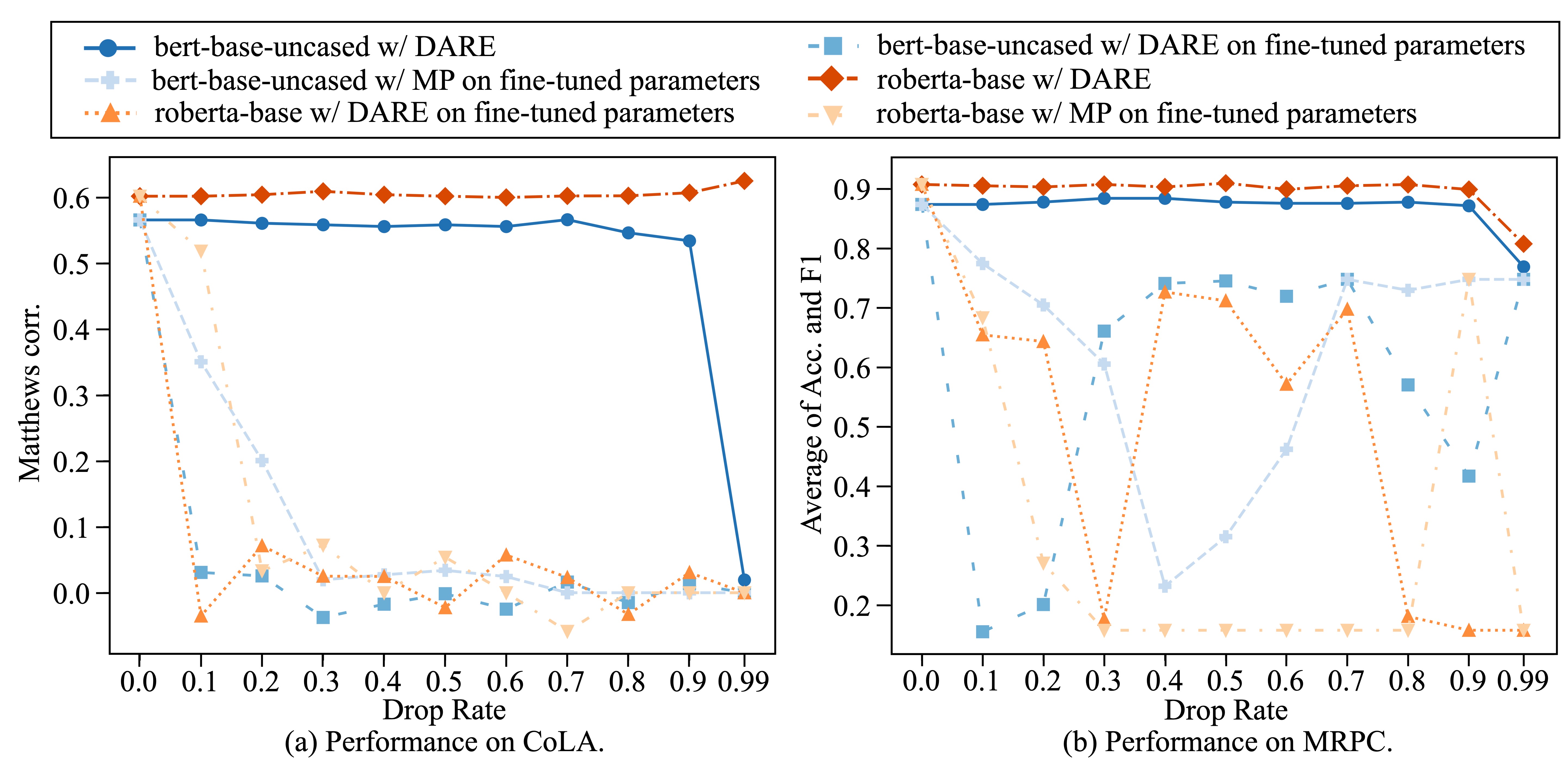}
    \caption{Results of DARE and MP by dropping fine-tuned parameters on CoLA and MRPC on encoder-based LMs.}
    \label{fig:plms_mask_finetuned_weight_comparison}
\end{figure}

As previous network pruning methods mainly operate on the fine-tuned instead of delta parameters, we also conduct experiments under this setting with both decoder- and . For decoder-based LMs, we find they perform badly when removing fine-tuned parameters even with 0.1 as the drop rate. Quantitatively, the performance sharply drops from 67.20 to 8.56 on AlpacaEval for WizardLM-13B, from 64.22/14.02 to 0.38/0.16 on GSM8K/MATH for WizardMath-13B, from 63.41/55.40 to 0.0/0.20 on HumanEval/MBPP for WizardCoder-Python-13B. Similar observations can also be found on MP or decoder-based LMs with 7B, 34B, or 70B sizes. Partial results on encoder-based LMs are shown in \figref{fig:plms_mask_finetuned_weight_comparison} and please see \figref{fig:plms_mask_finetuned_weight_comparison_all_results} in \secref{section-appendix-additional_results_plms_drop_fine_tuned_parameters} for additional results. 
We observe that directly eliminating the fine-tuned parameters by either DARE or MP would lead to worse performance on encoder-based LMs. The above results confirm that the knowledge is inherent in pre-trained LMs, and SFT is responsible for unlocking instead of introducing new capabilities. Moreover, decoder-based LMs are more susceptible than encoder-based LMs when removing fine-tuned parameters. This could be attributed to the fact that decoder-based LMs exhibit a higher degree of capability and have a stronger correlation with the fine-tuned parameters. Consequently, even the removal of a relatively small proportion of fine-tuned parameters can significantly degrade their performance.

\section{Conclusion}
\label{section-5}

In this work, we first discussed the extremely redundant properties of SFT delta parameters in LMs and proposed a simple approach DARE to effectively reduce the number of delta parameters needed for SFT without any data, retraining, or even GPUs. DARE can impressively drop 90\% or even 99\% SFT delta parameters without sacrificing much performance compared with using all SFT delta parameters. We further employed DARE as a versatile plug-and-play approach for existing model merging methods to merge multiple task-specific fine-tuned models into a single model with diverse abilities. Extensive experimental results on both encoder- and decoder-based LMs demonstrated the effectiveness of DARE in reducing SFT delta parameter redundancy and facilitating the model merging performance. We also provided a deeper analysis of why DARE works as well as the prerequisites for using DARE. We hope that our findings can advance the understanding of model alignment from the perspective of analyzing model parameters.

\section*{Impact Statement}
Recently, merging language models has become a promising research direction. Our work allows researchers to obtain a single model with diverse capabilities at a low cost. Thanks to our method, \textit{hundreds of models with different functionalities have been created on the Hugging Face community\footnote{\url{https://huggingface.co/models?other=arxiv:2311.03099}}. Several popular toolkits on the GitHub platform have also integrated our work, including huggingface/peft \footnote{\url{https://github.com/huggingface/peft}} and arcee-ai/mergekit\footnote{\url{https://github.com/arcee-ai/mergekit}}.} Even though this work has no direct social impacts, the potentially harmful information generated by LLMs (e.g., gender bias, racial discrimination) may still exist when using our approach. It is necessary to advocate for careful regulation by the communities as well as authorities on this matter.

\section*{Acknowledgements}
We would like to express our sincere gratitude to the anonymous reviewers for their insightful comments and suggestions, which have significantly enriched this paper. 

\bibliography{reference}
\bibliographystyle{icml2024}

%%%%%%%%%%%%%%%%%%%%%%%%%%%%%%%%%%%%%%%%%%%%%%%%%%%%%%%%%%%%%%%%%%%%%%%%%%%%%%%
%%%%%%%%%%%%%%%%%%%%%%%%%%%%%%%%%%%%%%%%%%%%%%%%%%%%%%%%%%%%%%%%%%%%%%%%%%%%%%%
% APPENDIX
%%%%%%%%%%%%%%%%%%%%%%%%%%%%%%%%%%%%%%%%%%%%%%%%%%%%%%%%%%%%%%%%%%%%%%%%%%%%%%%
%%%%%%%%%%%%%%%%%%%%%%%%%%%%%%%%%%%%%%%%%%%%%%%%%%%%%%%%%%%%%%%%%%%%%%%%%%%%%%%
\newpage
\appendix
\onecolumn
\label{section-appendix}

\section{Detailed Experimental Settings}
\subsection{Details of SFT and Pre-Trained Backbones of Decoder-based LMs}\label{section-appendix-llms_SFT_backbone_correspondences}
\tabref{tab:llms_SFT_backbone_correspondences} shows the versions and correspondences with pre-trained backbones of SFT decoder-based LMs.

\begin{table}[!htbp]
\centering
\caption{Versions and correspondences with pre-trained backbones of SFT decoder-based LMs.}
\label{tab:llms_SFT_backbone_correspondences}
% \resizebox{1.00\textwidth}{!}
% {
\setlength{\tabcolsep}{2.0mm}
{
\begin{tabular}{c|c|c}
\hline
Tasks                                   & SFT Decoder-based LMs                  & Pre-Trained Backbones       \\ \hline
\multirow{3}{*}{Instruction-following}  & WizardLM-7B\tablefootnote{\url{https://huggingface.co/WizardLM/WizardLM-7B-V1.0}}            & llama-7b\tablefootnote{\url{https://huggingface.co/decapoda-research/llama-7b-hf}}             \\
                                        & WizardLM-13B\tablefootnote{\url{https://huggingface.co/WizardLM/WizardLM-13B-V1.2}}           & Llama-2-13b\tablefootnote{\url{https://huggingface.co/meta-llama/Llama-2-13b-hf}}\saveFN{\llamaTwoThirteenBfn}          \\
                                        & WizardLM-70B\tablefootnote{\url{https://huggingface.co/WizardLM/WizardLM-70B-V1.0}}           & Llama-2-70b\tablefootnote{\url{https://huggingface.co/meta-llama/Llama-2-70b-hf}}\saveFN{\llamaTwoSeventyBfn}          \\ \hline
\multirow{3}{*}{Mathematical Reasoning} & WizardMath-7B\tablefootnote{\url{https://huggingface.co/WizardLM/WizardMath-7B-V1.0}}          & Llama-2-7b\tablefootnote{\url{https://huggingface.co/meta-llama/Llama-2-7b-hf}}           \\
                                        & WizardMath-13B\tablefootnote{\url{https://huggingface.co/WizardLM/WizardMath-13B-V1.0}}         & Llama-2-13b\useFN{\llamaTwoThirteenBfn}          \\
                                        & WizardMath-70B\tablefootnote{\url{https://huggingface.co/WizardLM/WizardMath-70B-V1.0}}         & Llama-2-70b\useFN{\llamaTwoSeventyBfn}          \\ \hline
\multirow{4}{*}{Code-generating}        & WizardCoder-Python-7B\tablefootnote{\url{https://huggingface.co/WizardLM/WizardCoder-Python-7B-V1.0}}  & CodeLlama-7b-Python\tablefootnote{\url{https://huggingface.co/codellama/CodeLlama-7b-Python-hf}}  \\
                                        & WizardCoder-Python-13B\tablefootnote{\url{https://huggingface.co/WizardLM/WizardCoder-Python-13B-V1.0}} & CodeLlama-13b-Python\tablefootnote{\url{https://huggingface.co/codellama/CodeLlama-13b-Python-hf}} \\
                                        & WizardCoder-Python-34B\tablefootnote{\url{https://huggingface.co/WizardLM/WizardCoder-Python-34B-V1.0}} & CodeLlama-34b-Python\tablefootnote{\url{https://huggingface.co/codellama/CodeLlama-34b-Python-hf}} \\
                                        & llama-2-13b-code-alpaca\tablefootnote{\url{https://huggingface.co/layoric/llama-2-13b-code-alpaca}}     & Llama-2-13b\useFN{\llamaTwoThirteenBfn}           \\ \hline
\end{tabular}
}
% }
\end{table}

\subsection{Learning Rate Configurations of Encoder-based LMs on GLUE}\label{section-appendix-plms_learning_rate_configuration}
The optimal settings of the learning rate of each fine-tuned encoder-based LM are presented in \tabref{tab:plms_learning_rate_configuration}.
\begin{table}[!htbp]
\centering
\caption{Configurations of learning rates of bert-base-uncased and roberta-base on GLUE.}
\label{tab:plms_learning_rate_configuration}
\setlength{\tabcolsep}{2.0mm}
{
\begin{tabular}{c|cccccccc}
\hline
Models    & CoLA & SST-2 & MRPC & STS-B & QQP & MNLI & QNLI & RTE \\ \hline
bert-base-uncased\tablefootnote{\url{https://huggingface.co/bert-base-uncased}}   & 5e-5   & 1e-5   & 5e-5  & 5e-5 & 1e-5  & 1e-5 & 1e-5        & 1e-5       \\
roberta-base\tablefootnote{\url{https://huggingface.co/roberta-base}}      & 1e-5   & 1e-5   & 5e-5  & 1e-5 & 1e-5  & 1e-5 & 1e-5        & 1e-5      \\ \hline
\end{tabular}
}
\end{table}

\subsection{Descriptions of Existing Model Merging Methods}\label{section-appendix-model_merging_methods_descriptions}
We experiment with five model merging methods:
\begin{itemize}
    \item  \textbf{Average Merging} simply averages the parameters of multiple models to get the merged model \cite{DBLP:conf/icml/WortsmanIGRLMNF22}.
    
    \item \textbf{Task Arithmetic} uses a scaling term to control the contributions between the pre-trained backbone and the models to be merged \cite{DBLP:conf/iclr/IlharcoRWSHF23}.

    \item \textbf{Fisher Merging} first estimates the importance of parameters by calculating the Fisher information matrix, and then fuses parameters based on their importance \cite{DBLP:conf/nips/MatenaR22}.
    
    \item \textbf{RegMean} recasts the model merging task as a linear regression problem and derives closed-form solutions to solve the problem \cite{DBLP:conf/iclr/Jin0P023}.

    \item \textbf{TIES-Merging} aims to address parameter conflicts in model merging. It first trims parameters with lower magnitudes, and then resolves sign disagreements. Parameters with consistent signs are finally merged \cite{DBLP:journals/corr/abs-2306-01708}. 
\end{itemize}

\subsection{Details of Grid Search on Hyperparameters of Model Merging Methods for Encoder-based LMs}\label{section-appendix-plms_grid_search_model_merging_hyperparameters}
\tabref{tab:plms_hyperparameter_searched_ranges_merging_methods} shows the searched ranges of model merging methods' hyperparameters for encoder-based LMs. For DARE, we search the drop rate $p$ in [0.1, 0.2, $\cdots$, 0.9] and select the optimal setting with the best performance. 
\begin{table}[!htbp]
\centering
\caption{Searched ranges of hyperparameters of model merging methods for encoder-based LMs.}
\label{tab:plms_hyperparameter_searched_ranges_merging_methods}
\setlength{\tabcolsep}{2.0mm}
{
\begin{tabular}{c|c}
\hline
Model Merging Methods & Search Ranges of Hyperparameters                                                                                                                                                                                 \\ \hline
Task Arithmetic       & \begin{tabular}[c]{@{}c@{}}scaling term to merge model parameters: {[}0.1, 0.3, 0.5, 0.7, 0.9, 1.0{]}\end{tabular}                                                                                            \\ \hline
Fisher Merging        & \begin{tabular}[c]{@{}c@{}}scaling term to merge model parameters: {[}0.1, 0.3, 0.5, 0.7, 0.9, 1.0{]}, \\ number of examples to compute Fisher information matrix: {[}256, 512, 1024, 2048{]}\end{tabular} \\ \hline
RegMean               & \begin{tabular}[c]{@{}c@{}}scaling term to reduce non-diagonal items: {[}0.1, 0.3, 0.5, 0.7, 0.9, 1.0{]},\\ number of examples to compute inner product matrices: {[}256, 512, 1024, 2048{]}\end{tabular}  \\ \hline
TIES-Merging          & \begin{tabular}[c]{@{}c@{}}scaling term to merge model parameters: {[}0.1, 0.3, 0.5, 0.7, 0.9, 1.0{]},\\ ratio to retain parameters with largest-magnitude values: {[}0.1, 0.2, 0.3{]}\end{tabular}         \\ \hline
\end{tabular}
}
\end{table}

\subsection{Details of Our Merged 7B LMs and the Open LLM Leaderboard}\label{section-appendix-details_7b_merged_model_leaderboard}
We offer two merged LMs with 7 billion parameters, namely supermario\_v1 and supermario\_v2. Specifically, we choose NeuralBeagle14-7B\footnote{\url{https://huggingface.co/mlabonne/NeuralBeagle14-7B}} and Turdus\footnote{\url{https://huggingface.co/udkai/Turdus}} to build supermario\_v1, where both of them all derived from Beagle14-7B\footnote{\url{https://huggingface.co/mlabonne/Beagle14-7B}}. We set the drop rate $p$ in DARE to 0.3, and merge NeuralBeagle14-7B and Turdus by Task Arithmetic with 0.8 as the scaling term. We select WildMarcoroni-Variant1-7B\footnote{\url{https://huggingface.co/BarryFutureman/WildMarcoroni-Variant1-7B}} and WestSeverus-7B-DPO-v2\footnote{\url{https://huggingface.co/FelixChao/WestSeverus-7B-DPO-v2}} to obtain supermario\_v2, where both of them adopt Mistral-7B-v0.1\footnote{\url{https://huggingface.co/mistralai/Mistral-7B-v0.1}} \cite{DBLP:journals/corr/abs-2310-06825} as the backbone. The drop rate $p$ in DARE is set to 0.5, and the scaling term in Task Arithmetic is also 0.5.

The Open LLM Leaderboard\footnote{\url{https://huggingface.co/spaces/HuggingFaceH4/open_llm_leaderboard}} is established to evaluate open-sourced LLMs based on Eleuther AI Language Model Evaluation Harness \cite{eval-harness}, which contains six benchmarks including AI2 Reasoning Challenge (ARC) \cite{DBLP:journals/corr/abs-1803-05457}, HellaSwag \cite{DBLP:conf/acl/ZellersHBFC19}, MMLU \cite{DBLP:conf/iclr/HendrycksBBZMSS21}, TruthfulQA \cite{DBLP:conf/acl/LinHE22}, Winogrande \cite{DBLP:conf/aaai/SakaguchiBBC20}, and GSM8K \cite{DBLP:journals/corr/abs-2110-14168}. The average score on the six datasets is used for ranking models on the leaderboard. We refer interested readers to the original papers for detailed information on the datasets. 

Note that the results of Turdus on Open LLM Leaderboard are not available and we instead report the performance of Beagle14-7B in \tabref{tab:llms_merging_open_llm_leaderboard_results}. Moreover, due to space limits, we use Hella., TQA, and Wino. as the abbreviations for HellaSwag, TruthfulQA, and Winogrande. WildMarcoroni-7B and WestSeverus-7B are the abbreviations for WildMarcoroni-Variant1-7B and WestSeverus-7B-DPO-v2.

\section{Additional Experimental Results}
\subsection{Additional Results of Delta Parameter Redundancy of Decoder-based LMs}\label{section-appendix-additional_results_llms_drop_rate_scaling_curve_AlpacaEval_MATH_MBPP}
\begin{figure}[!htbp]
    \centering
    \includegraphics[width=1.0\columnwidth]{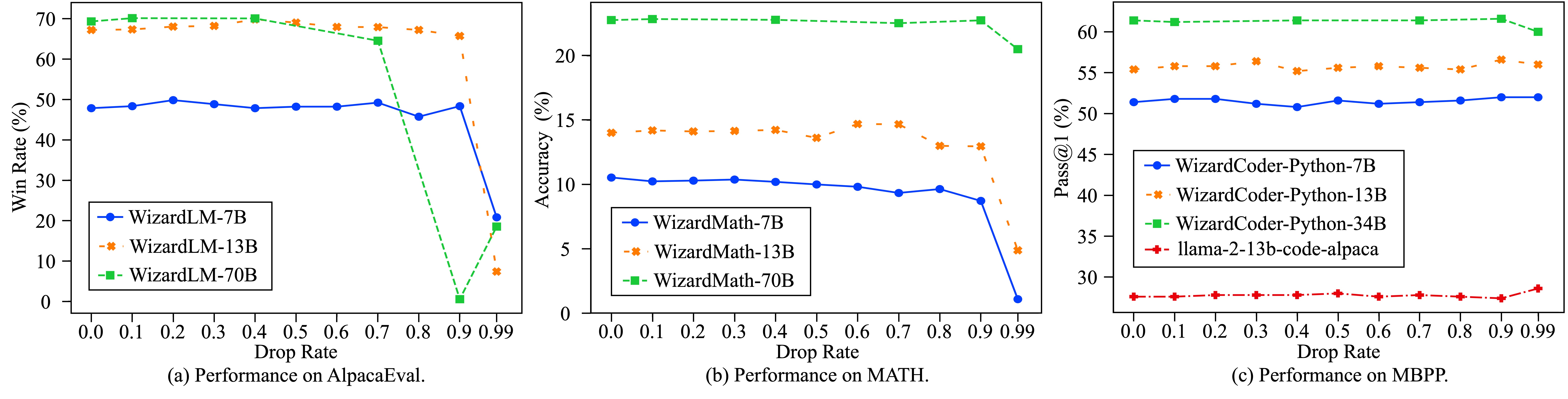}
    \caption{Performance of decoder-based LMs on AlpacaEval, MATH, and MBPP with various drop rates.}
    \label{fig:llms_drop_rate_scaling_curve_AlpacaEval_MATH_MBPP}
\end{figure}

\figref{fig:llms_drop_rate_scaling_curve_AlpacaEval_MATH_MBPP} shows results of decoder-based LMs on AlpacaEval, MATH, and MBPP with different drop rates. We notice that the performance of WizardLM-70B drastically declines on AlpacaEval when the drop rate is 0.9 (different from the observations of WizardMath-70B and WizardCoder-Python-34B). One possible reason is that the instruction-following task on AlpacaEval is harder and requires general abilities with more delta parameters via SFT, causing more obvious dependencies among parameters (especially on LMs with larger sizes). Therefore, when the ratio of dropped delta parameters reaches a relatively small value (e.g., 0.9 in this case), the dependent relationships among parameters are destroyed, leading to unsatisfactory performance.

\subsection{Additional Results of Merging Encoder-based LMs}\label{section-appendix-additional_results_lms_model_merging_DARE}
\begin{figure}[!htbp]
    \centering
    \includegraphics[width=1.00\columnwidth]{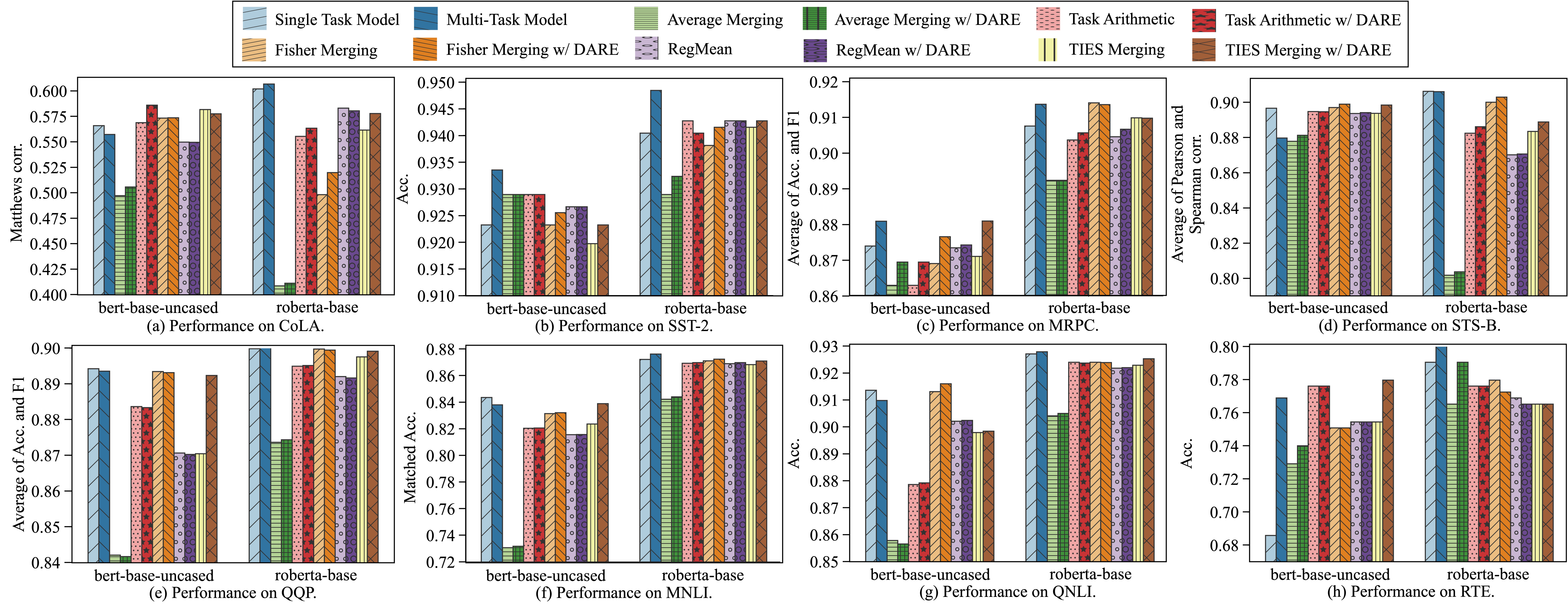}
    \caption{Performance of merging encoder-based LMs on GLUE.}
    \label{fig:plms_model_merging_all_results}
\end{figure}

\figref{fig:plms_model_merging_all_results} shows the performance of merging encoder-based LMs on GLUE.

\subsection{Additional Results of Comparisons between DARE and DropOnly}\label{section-appendix-additional_results_comparisons_DARE_with_DropOnly}
\begin{figure}[!htbp]
    \centering
    \includegraphics[width=1.00\columnwidth]{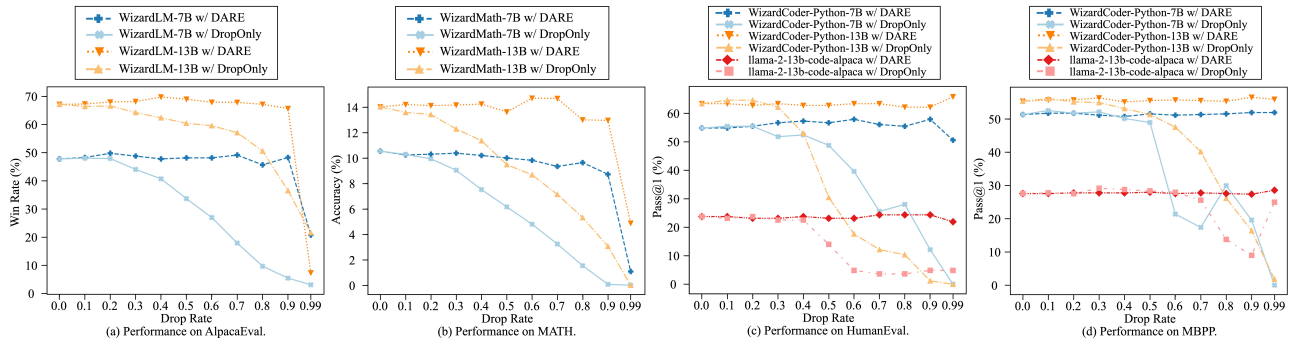}
    \caption{Comparing DARE and DropOnly on AlpacaEval, MATH, HumanEval, and MBPP on decoder-based LMs.}
    \label{fig:llms_rescale_comparison_AlpacaEval_MATH_HumanEval_MBPP}
\end{figure}

\begin{figure}[!htbp]
    \centering
    \includegraphics[width=1.00\columnwidth]{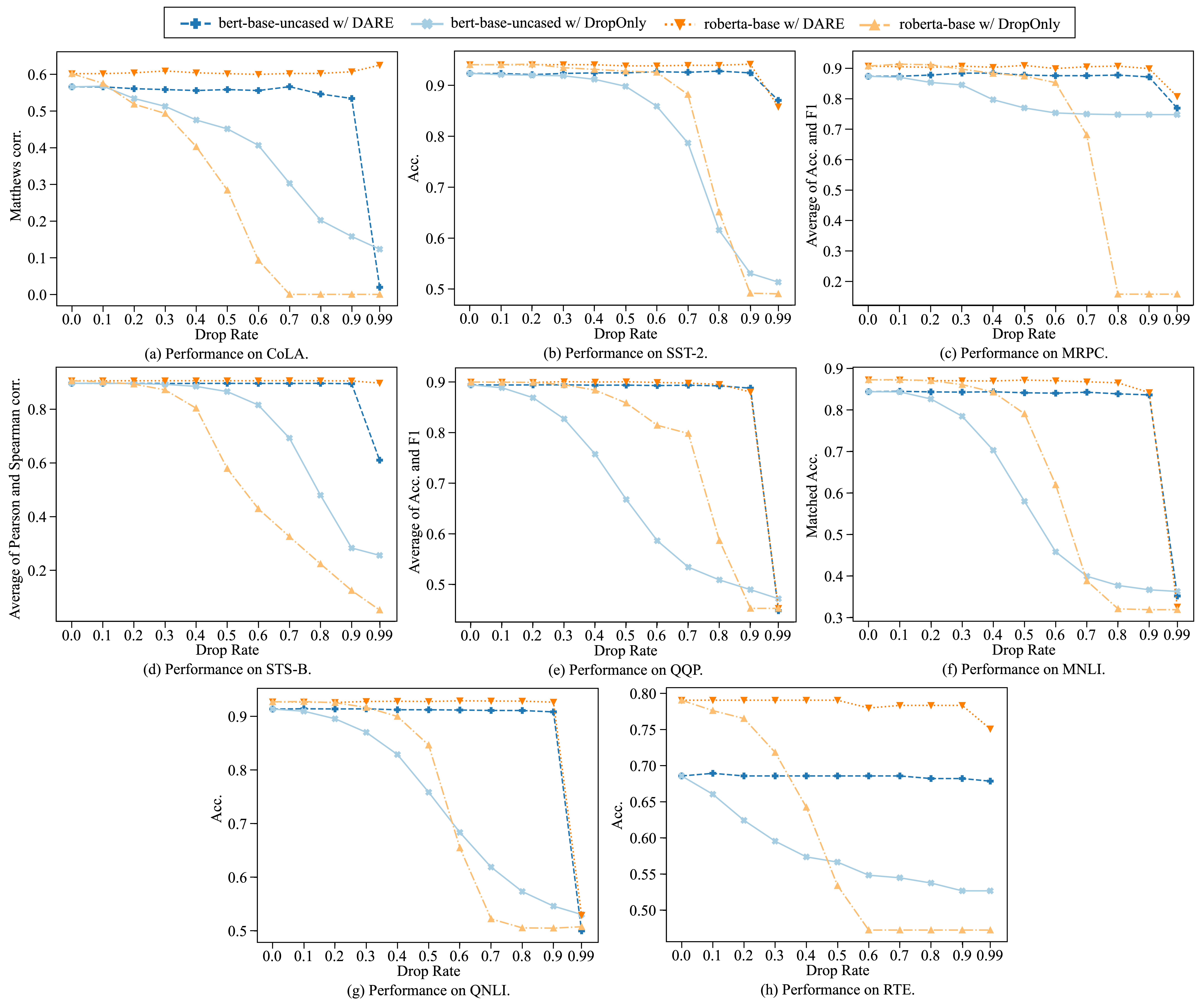}
    \caption{Comparisons between DARE and DropOnly on GLUE on encoder-based LMs.}
    \label{fig:plms_rescale_comparison_all_results}
\end{figure}

The comparison results between DARE and DropOnly on AlpacaEval, MATH, HumanEval, and MBPP on decoder-based LMs and all results on GLUE on encoder-based LMs are shown in \figref{fig:llms_rescale_comparison_AlpacaEval_MATH_HumanEval_MBPP} and \figref{fig:plms_rescale_comparison_all_results}, respectively.

\begin{figure}[!htbp]
    \centering
    \includegraphics[width=1.00\columnwidth]{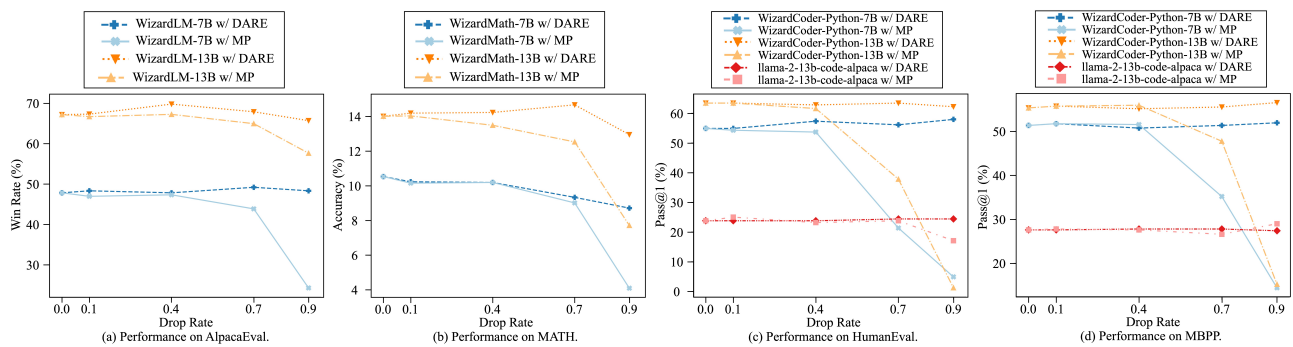}
    \caption{Comparisons between DARE and MP on AlpacaEval, MATH, HumanEval, and MBPP on decoder-based LMs.}
    \label{fig:llms_magnitude_pruning_comparison_AlpacaEval_MATH_HumanEval_MBPP}
\end{figure}

\subsection{Additional Results of Comparisons between DARE and MP}\label{section-appendix-additional_results_comparisons_DARE_with_MP}
\begin{figure}[!htbp]
    \centering
    \includegraphics[width=1.00\columnwidth]{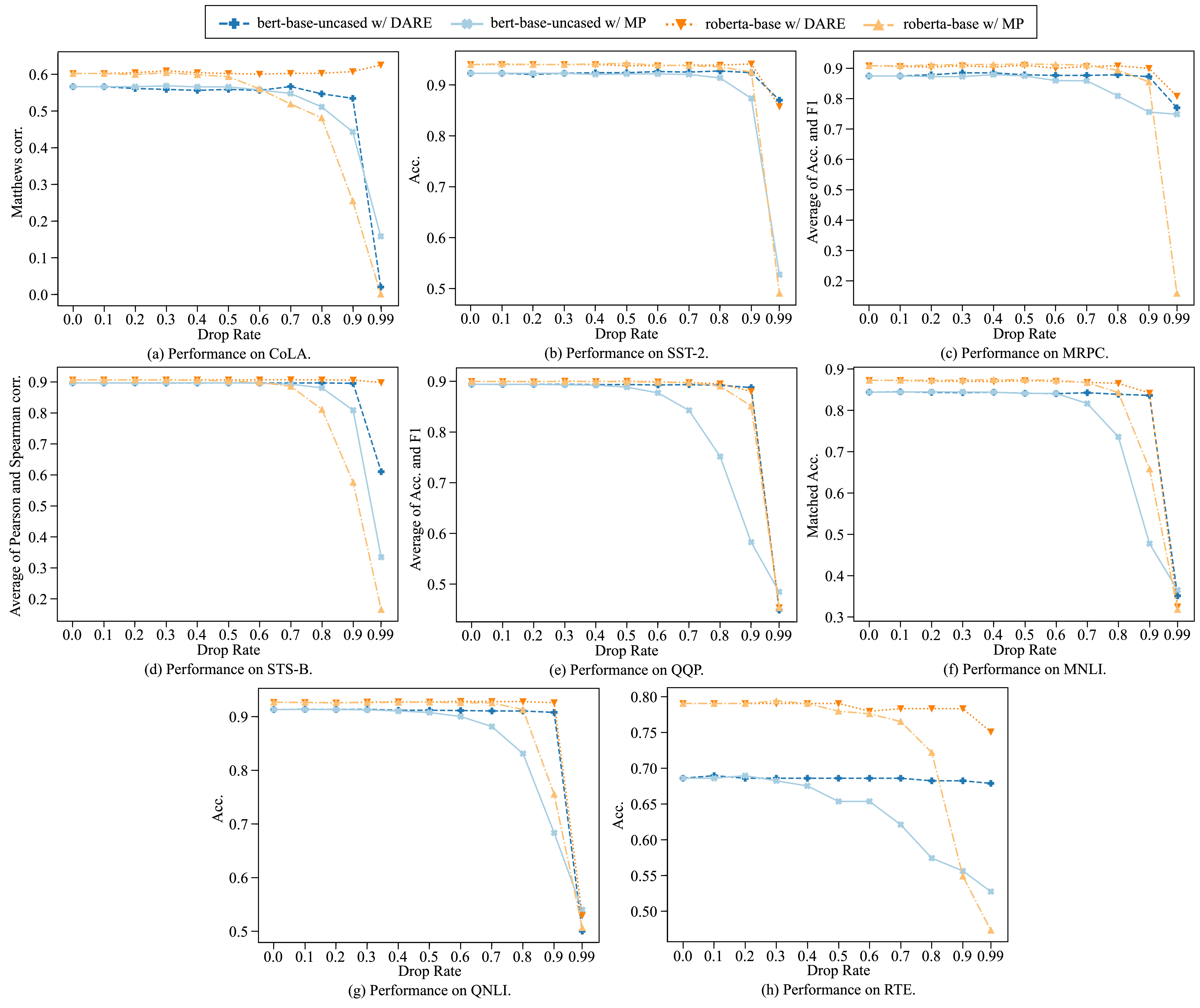}
    \caption{Comparisons between DARE and MP on GLUE on encoder-based LMs.}
    \label{fig:plms_magnitude_pruning_comparison_all_results}
\end{figure}

Comparisons between DARE and magnitude-based pruning on AlpacaEval, MATH, HumanEval, and MBPP on decoder-based LMs and all results on GLUE on encoder-based LMs are shown in \figref{fig:llms_magnitude_pruning_comparison_AlpacaEval_MATH_HumanEval_MBPP} and \figref{fig:plms_magnitude_pruning_comparison_all_results}, respectively.

\subsection{Ranges of SFT Delta Parameters of Decoder-based LMs and Encoder-based LMs}\label{section-appendix-additional_results_original_delta_parameter_ranges}
\begin{figure}[!htbp]
    \centering
    \includegraphics[width=1.00\columnwidth]{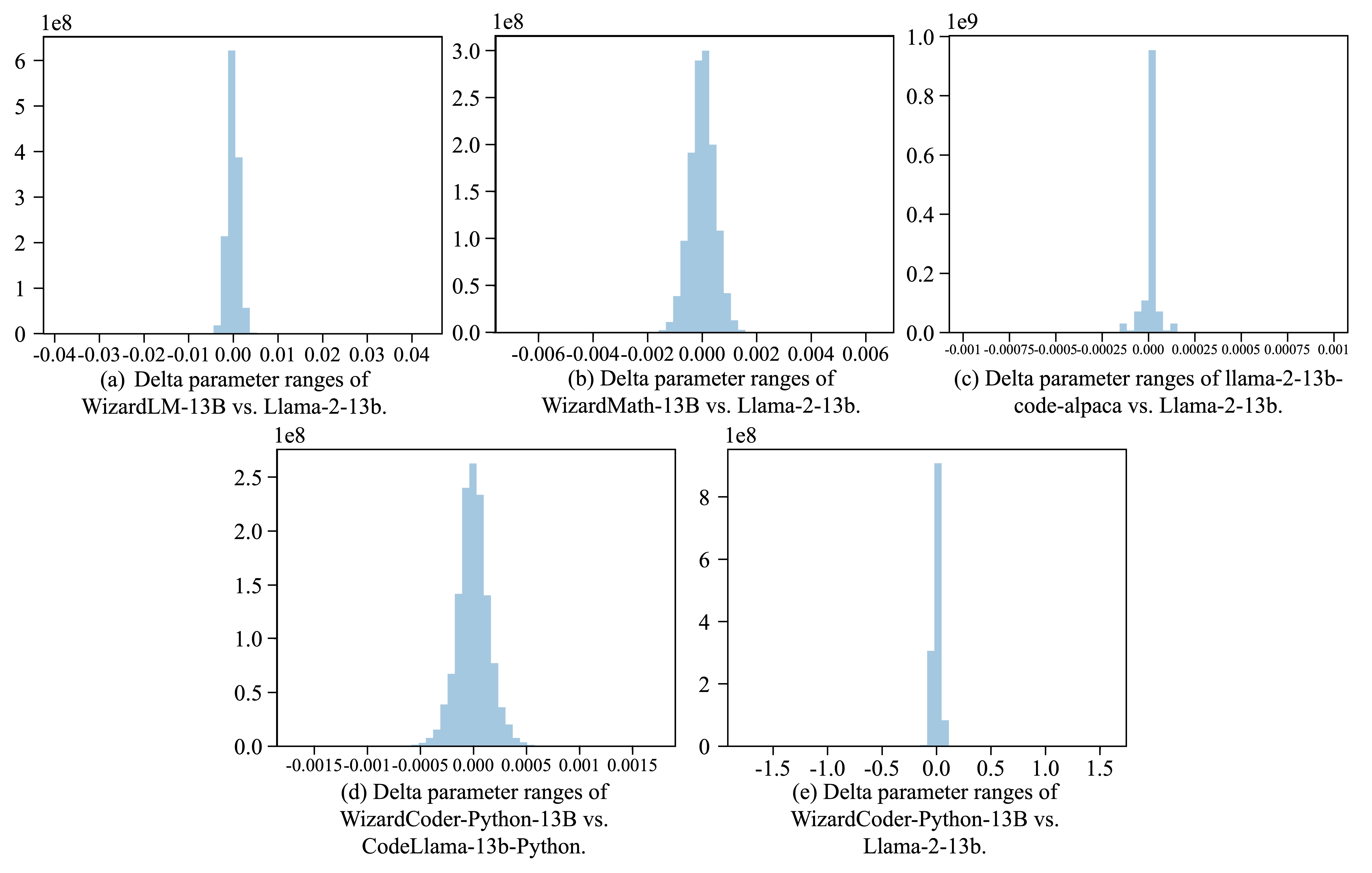}
    \caption{Delta parameter ranges of 13B decoder-based LMs vs. the pre-trained backbones.}
    \label{fig:llms_parameter_change_range}
\end{figure}

\begin{figure}[!htbp]
    \centering
    \includegraphics[width=1.00\columnwidth]{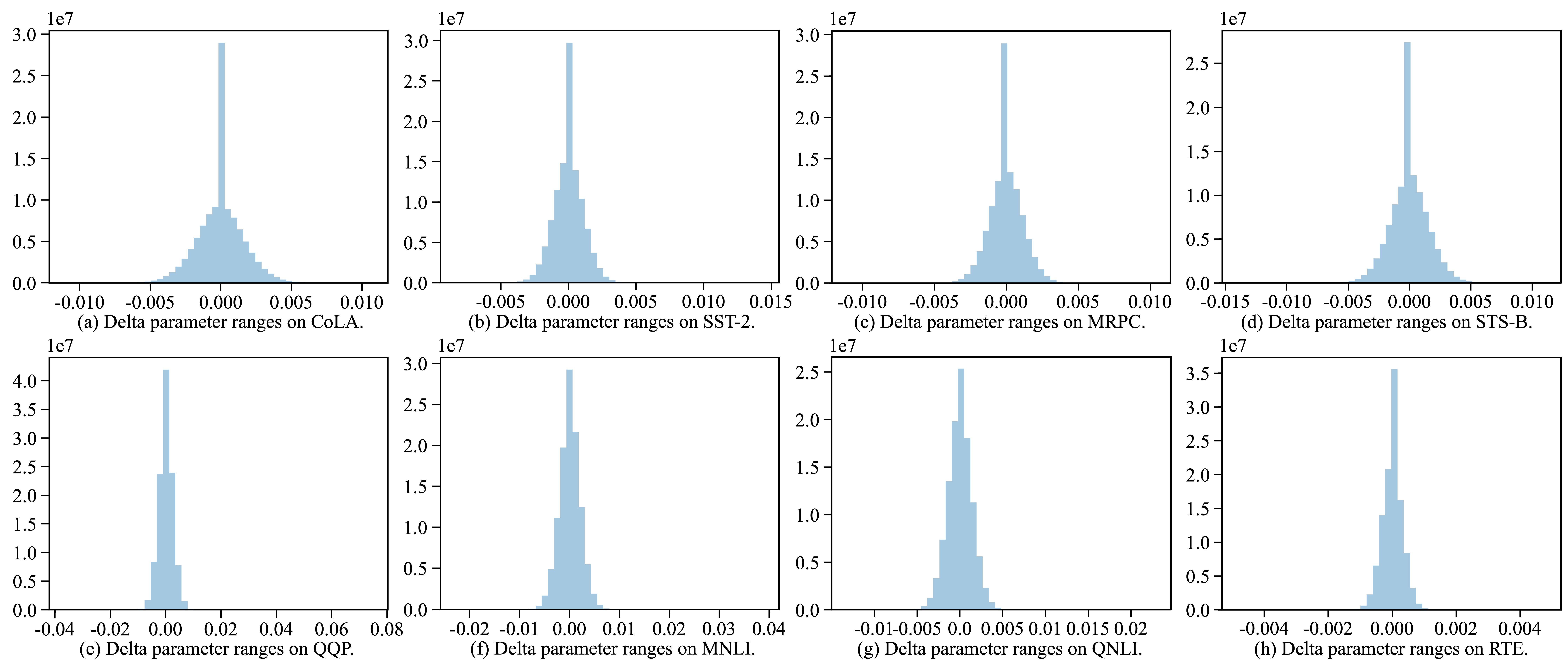}
    \caption{Delta parameter ranges of bert-base-uncased after SFT on GLUE.}
    \label{fig:bert_parameter_change_range}
\end{figure}

\begin{figure}[!htbp]
    \centering
    \includegraphics[width=1.00\columnwidth]{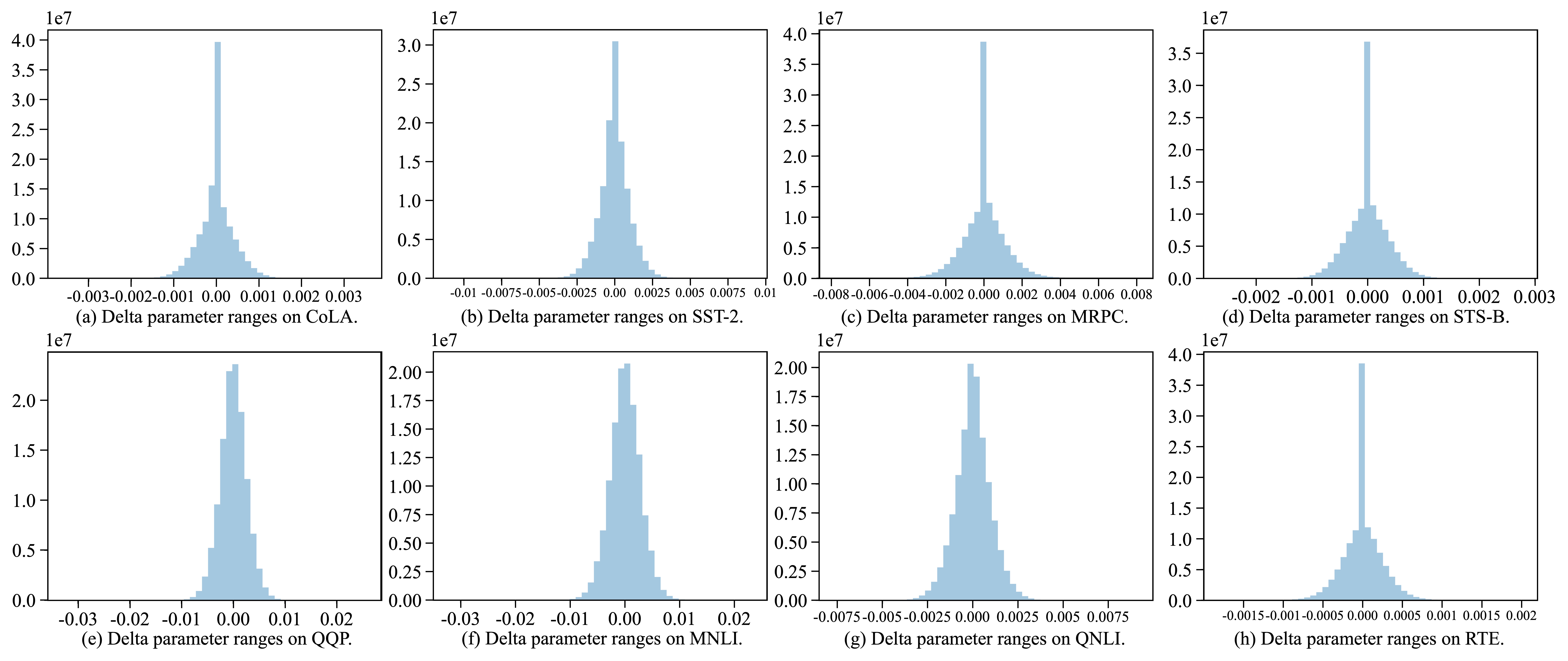}
    \caption{Delta parameter ranges of roberta-base after SFT on GLUE.}
    \label{fig:roberta_parameter_change_range}
\end{figure}

We show the SFT delta parameter ranges of decoder- and encoder-based LMs in \figref{fig:llms_parameter_change_range}, \figref{fig:bert_parameter_change_range} and \figref{fig:roberta_parameter_change_range}. Note that for decoder-based LMs, the results are obtained by randomly selecting 10\% delta parameters, whereas for encoder-based LMs, all delta parameters are included. We also provide the statistics on the percentiles of delta parameter ranges in \tabref{tab:statistics_parameters_changed_ranges}, which are derived by sorting the entire ranges and indexing at positions corresponding to 0, 10\%, 20\%, ..., 100\%.

\begin{sidewaystable}[!htbp]
\centering
\caption{Statistics about the deciles of delta parameter ranges of both decoder- and encoder-based LMs.}
\label{tab:statistics_parameters_changed_ranges}
\resizebox{0.95\textwidth}{!}
{
\setlength{\tabcolsep}{0.9mm}
{
\begin{tabular}{cc|ccccccccccc}
\hline
\multicolumn{2}{c|}{Models}                                                                                            & 0\% (min)         & 10\%        & 20\%        & 30\%        & 40\%        & 50\%       & 60\%       & 70\%       & 80\%       & 90\%       & 100\% (max)        \\ \hline
\multicolumn{2}{c|}{\begin{tabular}[c]{@{}c@{}}WizardLM-13B\\ vs. Llama-2-13b\end{tabular}}                    & -3.93e-02 & -0.16e-02 & -0.10e-02 & -0.06e-02 & -0.03e-02 &  0.00 &  0.03e-02 &  0.06e-02 & 0.10e-02 &  0.16e-02 &  4.81e-02     \\ \hline
\multicolumn{2}{c|}{\begin{tabular}[c]{@{}c@{}}WizardMath-13B\\ vs. Llama-2-13b\end{tabular}}                  & -0.69e-02 & -0.06e-02 & -0.04e-02 & -0.02e-02 & -0.01e-02 &  0.00 &  0.01e-02 &  0.02e-02 & 0.04e-02 &  0.06e-02 &  0.74e-02     \\ \hline
\multicolumn{2}{c|}{\begin{tabular}[c]{@{}c@{}}llama-2-13b-code-alpaca\\ vs. Llama-2-13b\end{tabular}}              & -8.42e-02 & -3.05e-05 &  0.00 &  0.00 &  0.00 & 0.00 &  0.00 &  0.00 &  0.00 &  3.05e-05 & 7.98e-02 \\ \hline
\multicolumn{2}{c|}{\begin{tabular}[c]{@{}c@{}}WizardCoder-Python-13B\\ vs. CodeLlama-13b-Python\end{tabular}} & -1.86e-03 & -1.81e-04 & -1.07e-04 & -6.87e-05 & -3.34e-05 & 0.00 &  3.34e-05 &  6.87e-05 &  1.07e-04 &  1.81e-04 & 1.82e-03 \\ \hline
\multicolumn{2}{c|}{\begin{tabular}[c]{@{}c@{}}WizardCoder-Python-13B\\ vs. Llama-2-13b\end{tabular}}          & -2.40 & -3.81e-02 & -2.47e-02 & -1.53e-02 & -7.31e-03 & 7.63e-06 &  7.32e-03 &  1.53e-02 &  2.47e-02 &  3.81e-02 & 2.40     \\ \hline
\multicolumn{1}{c|}{\multirow{8}{*}{bert-base-uncased}}                             & CoLA                             & -1.10e-02 & -1.99e-03 & -1.16e-03 & -5.70e-04 & -7.35e-05 & 3.08e-05 &  1.07e-04 &  5.74e-04 &  1.17e-03 &  1.99e-03 & 1.08e-02 \\
\multicolumn{1}{c|}{}                                                               & SST-2                            & -8.33e-03 & -1.33e-03 & -8.09e-04 & -4.29e-04 & -1.11e-04 & 4.81e-05 &  1.93e-04 &  4.42e-04 &  8.21e-04 &  1.34e-03 & 1.44e-02 \\
\multicolumn{1}{c|}{}                                                               & MRPC                             & -1.10e-02 & -1.41e-03 & -8.45e-04 & -4.49e-04 & -1.01e-04 & 8.55e-06 &  1.06e-04 &  4.54e-04 &  8.50e-04 &  1.41e-03 & 1.03e-02 \\
\multicolumn{1}{c|}{}                                                               & STS-B                            & -1.40e-02 & -1.88e-03 & -1.13e-03 & -5.93e-04 & -1.27e-04 & 2.11e-05 &  1.27e-04 &  5.93e-04 &  1.13e-03 &  1.88e-03 & 1.11e-02 \\
\multicolumn{1}{c|}{}                                                               & QQP                              & -3.66e-02 & -0.32e-02 & -0.19e-02 & -0.11e-02 & -0.04e-02 &  0.01e-02 &  0.06e-02 &  0.13e-02 & 0.21e-02 &  0.32e-02 &  7.47e-02     \\
\multicolumn{1}{c|}{}                                                               & MNLI                             & -2.28e-02 & -2.61e-03 & -1.63e-03 & -9.51e-04 & -3.90e-04 & 7.63e-05 &  4.81e-04 &  1.03e-03 &  1.70e-03 &  2.65e-03 & 3.87e-02 \\
\multicolumn{1}{c|}{}                                                               & QNLI                             & -1.32e-02 & -1.77e-03 & -1.11e-03 & -6.57e-04 & -2.76e-04 & 4.43e-05 &  3.31e-04 &  7.08e-04 &  1.16e-03 &  1.79e-03 & 2.29e-02 \\
\multicolumn{1}{c|}{}                                                               & RTE                              & -4.86e-03 & -3.94e-04 & -2.42e-04 & -1.31e-04 & -3.43e-05 & 2.12e-06 &  3.56e-05 &  1.33e-04 &  2.43e-04 &  3.95e-04 & 4.81e-03 \\ \hline
\multicolumn{1}{c|}{\multirow{8}{*}{roberta-base}}                                  & CoLA                             & -3.60e-03 & -4.94e-04 & -2.65e-04 & -1.01e-04 & -2.52e-05 & 1.81e-06 &  2.89e-05 &  1.01e-04 &  2.66e-04 &  4.97e-04 & 3.52e-03 \\
\multicolumn{1}{c|}{}                                                               & SST-2                            & -1.10e-02 & -1.18e-03 & -6.69e-04 & -3.43e-04 & -1.44e-04 & 1.04e-05 &  1.64e-04 &  3.81e-04 &  6.82e-04 &  1.19e-03 & 9.08e-03 \\
\multicolumn{1}{c|}{}                                                               & MRPC                             & -7.82e-03 & -1.26e-03 & -6.87e-04 & -2.95e-04 & -5.84e-05 & 2.92e-06 &  7.03e-05 &  2.94e-04 &  6.86e-04 &  1.26e-03 & 8.06e-03 \\
\multicolumn{1}{c|}{}                                                               & STS-B                            & -2.68e-03 & -4.45e-04 & -2.54e-04 & -1.11e-04 & -2.10e-05 & 1.22e-06 &  3.39e-05 &  1.12e-04 &  2.55e-04 &  4.46e-04 & 2.77e-03 \\
\multicolumn{1}{c|}{}                                                               & QQP                              & -3.29e-02 & -3.16e-03 & -2.01e-03 & -1.18e-03 & -5.33e-04 & 6.27e-05 &  6.51e-04 &  1.32e-03 &  2.14e-03 &  3.30e-03 & 2.56e-02 \\
\multicolumn{1}{c|}{}                                                               & MNLI                             & -3.22e-02 & -3.39e-03 & -2.17e-03 & -1.28e-03 & -5.78e-04 & 6.97e-05 &  7.09e-04 &  1.43e-03 &  2.31e-03 &  3.54e-03 & 2.31e-02 \\
\multicolumn{1}{c|}{}                                                               & QNLI                             & -7.69e-03 & -1.22e-03 & -7.51e-04 & -4.35e-04 & -1.83e-04 & 8.53e-06 &  2.10e-04 &  4.54e-04 &  7.67e-04 &  1.23e-03 & 9.11e-03 \\
\multicolumn{1}{c|}{}                                                               & RTE                              & -1.81e-03 & -2.90e-04 & -1.63e-04 & -7.32e-05 & -1.01e-05 & 3.86e-07 &  1.24e-05 &  7.38e-05 &  1.63e-04 &  2.91e-04 & 2.01e-03 \\ \hline
\end{tabular}
}
}
\end{sidewaystable}

\subsection{Additional Results of Dropping Fine-tuned Parameters on Encoder-based LMs}\label{section-appendix-additional_results_plms_drop_fine_tuned_parameters}
\begin{figure}[!htbp]
    \centering
    \includegraphics[width=1.00\columnwidth]{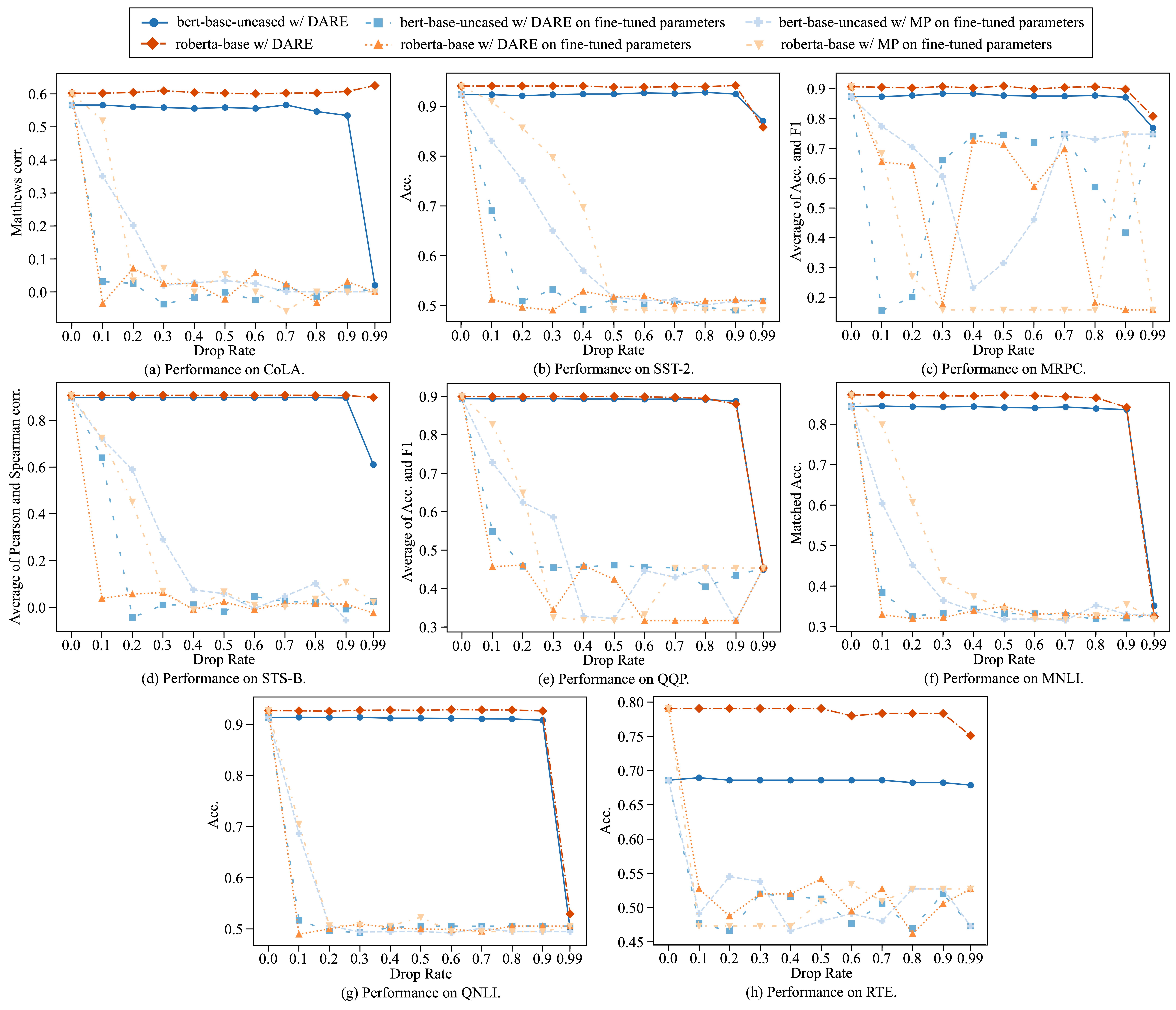}
    \caption{Performance of DARE and MP when dropping fine-tuned parameters on GLUE on encoder-based LMs.}
    \label{fig:plms_mask_finetuned_weight_comparison_all_results}
\end{figure}

\figref{fig:plms_mask_finetuned_weight_comparison_all_results} shows the results of removing fine-tuned parameters on GLUE on encoder-based LMs.

%%%%%%%%%%%%%%%%%%%%%%%%%%%%%%%%%%%%%%%%%%%%%%%%%%%%%%%%%%%%%%%%%%%%%%%%%%%%%%%
%%%%%%%%%%%%%%%%%%%%%%%%%%%%%%%%%%%%%%%%%%%%%%%%%%%%%%%%%%%%%%%%%%%%%%%%%%%%%%%

\end{document}